\documentclass[journal]{IEEEtran}
\usepackage{amsmath,amsfonts}
\usepackage{algorithmic}
\usepackage{algorithm}
\usepackage{array}
\usepackage[caption=false,font=footnotesize,labelfont=rm,textfont=rm]{subfig}
\usepackage{textcomp}
\usepackage{stfloats}
\usepackage{url}
\usepackage{verbatim}
\usepackage{graphicx}
\usepackage{cite}
\usepackage{multirow,booktabs,color,soul}
\usepackage{amsmath,rotating}
\usepackage{subfig} 
\usepackage{epsfig}
\usepackage{multirow,booktabs,color,soul,threeparttable,rotating}
\graphicspath{{Figure/}}
\usepackage{url}
\definecolor{hl}{rgb}{0.75,0.75,0.75}
\sethlcolor{hl}
\usepackage{hyperref}
\newcommand{\tabincell}[2]{\begin{tabular}{@{}#1@{}}#2\end{tabular}} 
\begin{document}
 
\title{Accurate and Efficient Event-based Semantic \\Segmentation Using Adaptive Spiking \\Encoder-Decoder Network }
% \title{A Sample Article Using IEEEtran.cls\\ for IEEE Journals and Transactions}
\author{IEEE Publication Technology,~\IEEEmembership{Staff,~IEEE,}
        % <-this % stops a space
\thanks{This paper was produced by the IEEE Publication Technology Group. They are in Piscataway, NJ.}% <-this % stops a space
\thanks{Manuscript received April 19, 2021; revised August 16, 2021.}}

% The paper headers
% \markboth{Journal of \LaTeX\ Class Files,~Vol.~14, No.~8, August~2021}%
% \markboth{IEEE Transactions on Neural Networks and Learning Systems,~Vol.~xx, No.~xx, xxxx}%
% % IEEE Transactions on Neural Networks and Learning Systems,~Vol.~xx, No.~xx, xxxx
% {Rui Zhang \MakeLowercase{\textit{et al.}}: Accurate and Efficient Event-based Semantic Segmentation with Adaptive Spiking Encoder-Decoder Network}

% \IEEEpubid{0000--0000/00\$00.00~\copyright~2021 IEEE}
% Remember, if you use this you must call \IEEEpubidadjcol in the second
% column for its text to clear the IEEEpubid mark.

\author{Rui Zhang,
        Luziwei Leng~\IEEEmembership{Member, IEEE},
        Kaiwei Che,
        Hu Zhang,
        Jie Cheng,
        Qinghai Guo, \\
        Jianxing Liao, 
        and~Ran Cheng~\IEEEmembership{Senior Member, IEEE}% <-this % stops a space 
        \thanks{Rui Zhang, Hu Zhang and Ran Cheng are with the Department of Computer Science and Engineering, Southern University of Science and Technology, Shenzhen 518055, China. Kaiwei Che is with the Department of Electrical and Electronic Engineering, Southern University of Science and Technology, Shenzhen 518055, China.
                (e-mail: ranchengcn@gmail.com). Rui Zhang did this work during her internship in ACSLab, Huawei Technologies Co., Ltd.
                }
        \thanks{Luziwei Leng, Jie Cheng, Qinghai Guo and Jianxing Liao are with the Advanced Computing and Storage Lab, Huawei Technologies Co., Ltd., Shenzhen 518055, China. (e-mail: lengluziwei@huawei.com) (\emph{Corresponding authors: Luziwei Leng and Ran Cheng.})}
}

\maketitle

\begin{abstract}
Spiking neural networks (SNNs), known for their low-power, event-driven computation and intrinsic temporal dynamics, are emerging as promising solutions for processing dynamic, asynchronous signals from event-based sensors. Despite their potential, SNNs face challenges in training and architectural design, resulting in limited performance in challenging event-based dense prediction tasks compared to artificial neural networks (ANNs). 
In this work, we develop an efficient spiking encoder-decoder network (SpikingEDN) for large-scale event-based semantic segmentation tasks. 
To enhance the learning efficiency from dynamic event streams, we harness the adaptive threshold which improves network accuracy, sparsity and robustness in streaming inference. 
Moreover, we develop a dual-path Spiking Spatially-Adaptive Modulation module, which is specifically tailored to enhance the representation of sparse events and multi-modal inputs, thereby considerably improving network performance. Our SpikingEDN attains a mean intersection over union (MIoU) of 72.57\% on the DDD17 dataset and 58.32\% on the larger DSEC-Semantic dataset, showing competitive results to the state-of-the-art ANNs while requiring substantially fewer computational resources. Our results shed light on the untapped potential of SNNs in event-based vision applications. The source code will be made publicly available.
\end{abstract}
% !!! recheck citations, 缩写重复解释


\IEEEpeerreviewmaketitle
\section{Introduction}\label{sec:introduction}

\IEEEPARstart{D}{erived} from biological neurons, spiking neuron models are characterized by their event-driven nature and inherent temporal dynamics. 
By processing information through discrete spikes rather than real numbers, particularly when implemented on neuromorphic chips with high efficiency, spiking neural networks (SNNs) consume significantly less energy than traditional artificial neural networks (ANNs) \cite{ding2024enhancing}. 
These capabilities not only make SNNs highly effective in safety-critical applications like autonomous driving and robotic control \cite{shen2024efficient}, but also enable them to handle complex spatio-temporal dynamics in continual learning scenarios \cite{shen2021hybridsnn}, allowing adaptation to new information without retraining and offering significant advantages in dynamic environments.
The sparse and asynchronous computation nature also make SNNs highly suitable for processing dynamic signals from retina-inspired, event-based sensors \cite{serrano2013128,brandli2014240,son20174}. 
Event data from neuromorphic vision sensors like event-based sensors excel with high temporal resolution and dynamic range, allowing for precise, low-power capture of rapid motions, ideal for detailed temporal tasks such as rapid image reconstruction and deblurring \cite{chen2024enhancing}. 
Unlike traditional ANNs that perform dense and static computations, SNNs process information dynamically and sparsely, aligning well with the event-driven nature of the data. This dynamic processing can potentially reduce redundancy and improve the system's ability to handle dynamic scenarios efficiently. 
The integration of event-based sensors with SNNs could potentially yield exceptionally fast and energy-efficient neuromorphic systems \cite{roy2019towards, davies2021advancing, merolla2014million, furber2014spinnaker, kungl2019accelerated, frenkel2022reckon, zhang2023automotive}.
These systems are particularly suited for edge devices where power resources are limited or in high-speed applications such as vehicles and drones. Particularly, in the context of event-based semantic segmentation\cite{alonso2019ev,sun2022ess,yang2023event}, this combination facilitates quick contextual comprehension, a task demanding dense output from inputs that are both sparse and dynamic, thus presenting a unique challenge.
% }

Presently, most leading event-based vision research predominantly uses ANNs, which rely on dense, frame-based computation \cite{ahmed2021deep,zhang2022discrete,alonso2019ev,sun2022ess, Luo_2023_ICCV, Ponghiran_2023_ICCV, Liu_2023_ICCV, Su_2023_ICCV}. A significant reason for this is the complexity of training SNNs using gradient-based methods compared to ANNs. In SNNs, the membrane potential of a neuron undergoes continuous evolution and emits a discrete spike when the potential exceeds a threshold. This spiking process, being discontinuous, is inherently incompatible with traditional gradient-based backpropagation, which requires continuous differentiable variables. The surrogate gradient approach addresses this challenge by replacing the non-differentiable function with continuous, smooth approximations \cite{zenke2018superspike,wu2018spatio,neftci2019surrogate}. By adopting this approach and implementing advanced learning algorithms, SNNs have made significant strides in classification tasks on various benchmark image and event-based datasets, achieving accuracy levels on par with ANNs \cite{rathi2021diet,zheng2021going,li2021differentiable,fang2021deep,deng2022temporal}.

However, SNNs have yet to demonstrate competitive capabilities in more complex vision tasks like dense prediction, where the network architecture poses a significant challenge. Unlike in simpler classification tasks, the sophisticated structures common in ANNs, featuring a variety of variations, are not easily adaptable to SNNs. Designing architectures for SNNs is constrained by the complexities associated with their training and the extended duration required, often resulting in simplistic network designs that deliver suboptimal performance \cite{hagenaars2021self,kim2021beyond,zhu2022event}. Additionally, the incompatibility of advanced deep learning techniques, such as attention mechanisms or normalization processes \cite{vaswani2017attention,ba2016layer,ulyanov2016instance}, with the multiplication-free inference (MFI) characteristic of spike-based computation, limits the integration of these powerful operations in SNNs. 
This limitation presents a significant obstacle in employing SNNs for challenging vision tasks, particularly when compared to the capabilities of ANNs.

% \textcolor{blue}{
In this study, we develop an efficient spiking encoder-decoder network (SpikingEDN) for event-based semantic segmentation across two large-scale datasets. To our knowledge, this marks the first instance where SNNs match the performance of sophisticated ANNs in these tasks. Our contributions are summarized as follows:
\begin{itemize}
\item We investigate the role of adaptive threshold in event encoding, and demonstrate that this inherent mechanism can improve network sparsity, accuracy and robustness, particularly in streaming inference.
\item To fully exploit the capabilities of SNNs in event-based semantic segmentation, we design an efficient spiking encoder-decoder network (i.e., SpikingEDN), which incorporates recent advancements in architecture search for SNNs \cite{liu2019auto, che2022differentiable}.
\item To augment the representation of sparse events and integrate multi-modal inputs with gray-scale images, we introduce an MFI-compatible, dual-path Spiking Spatially-Adaptive Modulation (SSAM) module. 
\item Our SpikingEDN achieves a mean intersection over union (MIoU) of 72.57\% on the DDD17 dataset \cite{binas2017ddd17}, surpassing directly trained SNNs by 37\% and ANNs by 4\% in terms of MIoU. On the high-resolution DSEC dataset \cite{sun2022ess,gehrig2021dsec}, our network attains a 58.32\% MIoU, demonstrating competitive performance with state-of-the-art ANN-based methods utilizing transfer learning. Additional evaluations on streaming inference, operational numbers, and potential energy costs confirm the robustness and significantly greater power efficiency of our networks compared to ANNs.
\end{itemize}
% }

The structure of this paper is organized as follows.
Section \ref{sec:background} provides essential background information, encompassing event-based semantic segmentation, the application of SNNs in dense prediction tasks, and adaptive threshold neurons.
Section \ref{sec:method} delves into a detailed examination of the spiking neuron with adaptive threshold, followed by an introduction of the  architecture of our SpikingEDN and the implementation of the SSAM module.
In Section \ref{sec:result}, we present extensive experimental studies conducted on the DDD17 and DSEC-Semantic datasets. This includes an in-depth study of adaptive thresholds, ablation studies on architectures, an analysis of network sparsity, evaluations of random seed variations, and tests incorporating gray-scale images.
Section \ref{sec:conclusion} concludes the paper by summarizing our key findings and providing final thoughts.

\section{Background}\label{sec:background}

%This section provides the necessary background on various topics that underpin our research. We begin by discussing the concept of Event-based Semantic Segmentation (EbSS), highlighting its potential and current advancements. We then delve into the role of SNNs in dense prediction tasks, exploring recent developments and challenges in this area. Lastly, we focus on the Adaptive Threshold Neuron, its previous applications, and how we propose to leverage its potential in our work.

\subsection{Event-based Semantic Segmentation (EbSS)}
Event-based cameras, notable for their exceptional temporal resolution (1us) and dynamic range (120dB) compared to traditional frame-based cameras, are increasingly favored for high-speed applications on edge platforms such as vehicles and drones. 
The realm of EbSS, while still in its nascent stages with a limited number of existing studies, is experiencing rapid growth in interest. EV-SegNet \cite{alonso2019ev} initially set a benchmark for EbSS on the expansive DDD17 dataset \cite{binas2017ddd17}. 
By adopting an Xception-based \cite{chollet2017xception} encoder-decoder structure, EV-SegNet introduced an event data representation that captured both event histograms and temporal distributions.
Subsequent research, which incorporates transfer learning methodologies \cite{wang2021evdistill, sun2022ess, yang2023event, xie2024cross}, has leveraged insights from high-quality image datasets.
These studies have achieved semantic segmentation on unlabeled event data via unsupervised domain adaptation, outperforming supervised methods. Nonetheless, the knowledge distillation and pre-training approaches they employed added to the computational demands. 
The study in \cite{kim2021beyond} marked the first attempt to directly train SNNs for EbSS using the DDD17 dataset. This research explored SNNs adapted from established ANN architectures, such as DeepLab \cite{chen2017deeplab} and Fully-Convolutional Networks (FCN) 
% \cite{chen2014semantic,long2015fully}. 
\cite{long2015fully}. However, despite these efforts, their models fell short of the accuracy achieved by leading ANNs and required an extensive number of time-steps to inference. A recent work \cite{das2024halsie} further explored hybrid SNN-ANN architecture to take advantages of both domains.

\subsection{SNNs for Dense Prediction}

Recent advancements in surrogate gradient methods \cite{shrestha2018slayer,wu2019direct, wozniak2020deep, li2021differentiable,fang2021deep,deng2022temporal, dampfhoffer2023backpropagation, zhou2022spikformer} have been instrumental in elevating SNNs to high levels of accuracy on benchmark image and event-based classification datasets, bringing them into close competition with ANNs.

In the realm of dense prediction tasks, such as event-based optical flow estimation \cite{hagenaars2021self}, stereo matching \cite{ranccon2021stereospike}, and video reconstruction \cite{zhu2022event}, SNNs have also been gaining traction. Despite these efforts, the accuracy of these SNNs, often constrained by traditional ANN architectures or simplistic custom-designed networks, tends to lag behind that of state-of-the-art ANNs. The importance of effective layer-level dimension variation is particularly pronounced in dense prediction tasks. In response to this challenge, recently, a spike-based differentiable hierarchical search method has been proposed \cite{che2022differentiable}. Drawing inspiration from differentiable architecture search techniques \cite{liu2018darts,liu2019auto}, this method shows promising results in optimizing SNNs for tasks like event-based deep stereo.

\subsection{Adaptive Threshold Neuron}
The adaptive threshold concept has traditionally been employed in recurrent SNNs to facilitate long-term memory, with its parameter typically set as non-trainable \cite{bellec2018long, bellec2020solution}. The study by \cite{hagenaars2021self} extended the use of the Adaptive Leaky Integrate-and-Fire (ALIF) neuron throughout the network for event-based optical flow estimation. However, this approach resulted in performance that was somewhat inferior to that achieved using standard LIF neurons. Further, \cite{frenkel2022reckon} demonstrated that LIF neurons with an extended time constant could match the performance of ALIF neurons in regression tasks. In our study, we focus on employing the adaptive threshold solely in the first layer and investigate
its short-term modulation functionality for event encoding in dense prediction tasks

\section{Methodology} \label{sec:method}

This section begins with an in-depth examination of event encoding using an adaptive threshold neuron. Subsequently, we describe the implementation of a hierarchical search method for developing SpikingEDN. Finally, we elaborate the the tailored MFI-compatible and dual-path SSAM module. 

\subsection{Event Encoding with Adaptive Threshold Neuron}
Event-based cameras distinctively capture intensity changes at each pixel. An event is represented as a tuple $(x, y, t, p)$, where $x$ and $y$ denote pixel coordinates, $t$ is the event's timestamp, and $p$ represents polarity, indicating an increase or decrease in brightness beyond a set threshold. To preserve the temporal aspect of these event streams, we utilize a Stacking Based on Time (SBT) approach \cite{wang2019event}. This method groups events into short temporal windows.

Throughout a specified duration $\Delta t$, events are compiled into $n$ sequential frames. The value at each pixel in frame $i$ is determined by summing the polarity of events:
\begin{equation}
P_i(x,y) = \sum_{t\in T} p(x,y,t),
\label{eq:sbt}
\end{equation}
where $P$ is the cumulative pixel value at $(x,y)$, $t$ is the timestamp, $p$ is the event's polarity, and $T \in [\frac{(i-1)\Delta t }{n},\frac{i\Delta t}{n}]$ represents the time window for event accumulation in each frame. The network processes these $n$ frames per stack, treating each as an individual input channel.

Although event-based cameras capture motion effectively, images provide a more comprehensive texture detail. In cases where enhanced input is beneficial, we merge images with event frames, integrating this combined data as an additional channel.
Automotive environments often feature scenarios where objects moving at various speeds produce event signals with significant density variations. Traditional ANNs, typically optimized for constant data rate images, struggle in these dynamic settings. While adaptive sampling methods that iteratively count events to form a frame can be effective, they tend to introduce extra computational demands. To address these challenges, we exploit the rich temporal dynamics inherent to spiking neurons, proposing innovative solutions for these dynamic scenes.

Drawing inspiration from the adaptive threshold mechanism \cite{bellec2018long}, we introduce the Adaptive Iterative Leaky-Integrate and Fire (AiLIF) neuron model, described as follows:
\begin{equation}
\begin{split}
u_i^{t, n} &= \tau u_i^{t-1,n}(1-y_i^{t-1,n}) + I_i^{t,n}, \\
y^{t} &= H(u^{t}-A^{t}), \\
A^{t} &= u_{\mathrm{th}}+\beta a^{t}, \\
a^{t} &= \tau_{a} a^{t-1}+y^{t-1}. \\
\end{split}
\label{eq2}
\end{equation}
In this model, $u_i^{t,n}$ represents the membrane potential of neuron $i$ in layer $n$ at time $t$, with $\tau$ as the membrane time constant. $y$ denotes the output spike and $I$ denotes the input current, defined as $I_i^{t,n} = \sum_{j} w_{ij} y^{t, n-1}_j$ where $w$ is the synaptic weight. A neuron fires a spike ($y = 1$) when its membrane potential surpasses a dynamic threshold $A^{t}$; otherwise, it remains inactive ($y = 0$). The Heaviside step function, $H$, used to determine the firing condition, is expressed as:
% \textcolor{blue}{
\begin{equation}
H(u^{t}-A^{t}) = \begin{cases}
0, & \text{if } u^{t}-A^{t} < 0, \\
1, & \text{if } u^{t}-A^{t} \geq 0.
\end{cases}
\end{equation}
% }

% \textcolor{blue}{
The time-varying threshold at time $t$ is represented by $A^t$, with $a^t$ being the cumulative threshold increment that adjusts based on the neuron's spiking history. The parameter $\beta$ serves as a scaling factor, while $\tau_a$ denotes a decay factor ($\tau_a \in (0,1)$) of $a^t$. This adjustable threshold, $A^{t}$, regulates the spiking rate, increasing with denser inputs to inhibit neuron firing, and decreasing otherwise, leading to a self-adaptive spiking behavior. It is noteworthy that our adaptive threshold mechanism differs slightly from that in \cite{bellec2018long}, primarily in the simplification of the decay factor $\tau_a$ and the omission of the weighting factor for spikes. This approach is more akin to the one used in \cite{bellec2020solution}, but here it is applied to an iterative LIF neuron with a hard reset mechanism.
% }

According to Eq. \ref{eq2}, in cases of prolonged inactivity, $a^{t}$ approaches zero as $t$ increases indefinitely, setting the lower bound of $A^t$ to the base threshold $u_{\mathrm{th}}$. Conversely, in a scenario of continuous activity, starting with $y^0=1$ and $a^0=0$, the expression for $a^{t}$ becomes
\begin{equation}
a^{t} = \sum_{i=1}^{t} \tau_{a}^{i-1},
\label{eq:eqac}
\end{equation}
and its upper bound in the long term equals $1/(1-\tau_{a})$. Thus, the adjustable threshold $A^t$ varies within the range $[u_{\mathrm{th}}, u_{\mathrm{th}}+\beta/(1-\tau_{a})]$.

In our experiments, we set the threshold $u_{th}$ to 0.5, the membrane time constant $\tau$ to 0.2, and treat $\tau_a$ as a trainable parameter, drawing inspiration from \cite{fang2021incorporating, zhang2022discrete}. Although it has been shown that the application of ALIF neuron throughout the entire network resulted in less optimal outcomes compared to the exclusive use of LIF neurons \cite{hagenaars2021self}, our hypothesis is that excessive flexibility might have hindered training precision, as evidenced in our latter study (refer to Section \ref{subsubsec:ablation}). To counteract this, we incorporate the AiLIF neuron only in the first layer for event encoding, while standard iterative LIF neurons, with $\beta$ set to 0 in Eq. (\ref{eq2}), are used in the subsequent layers. For training the SNN, we utilize the spatio-temporal backpropagation algorithm \cite{wu2018spatio} and Dspike \cite{li2021differentiable} as the surrogate gradient function.

\subsection{SSAM Modulation}
\begin{figure}[t]
\includegraphics[width=0.5\textwidth]{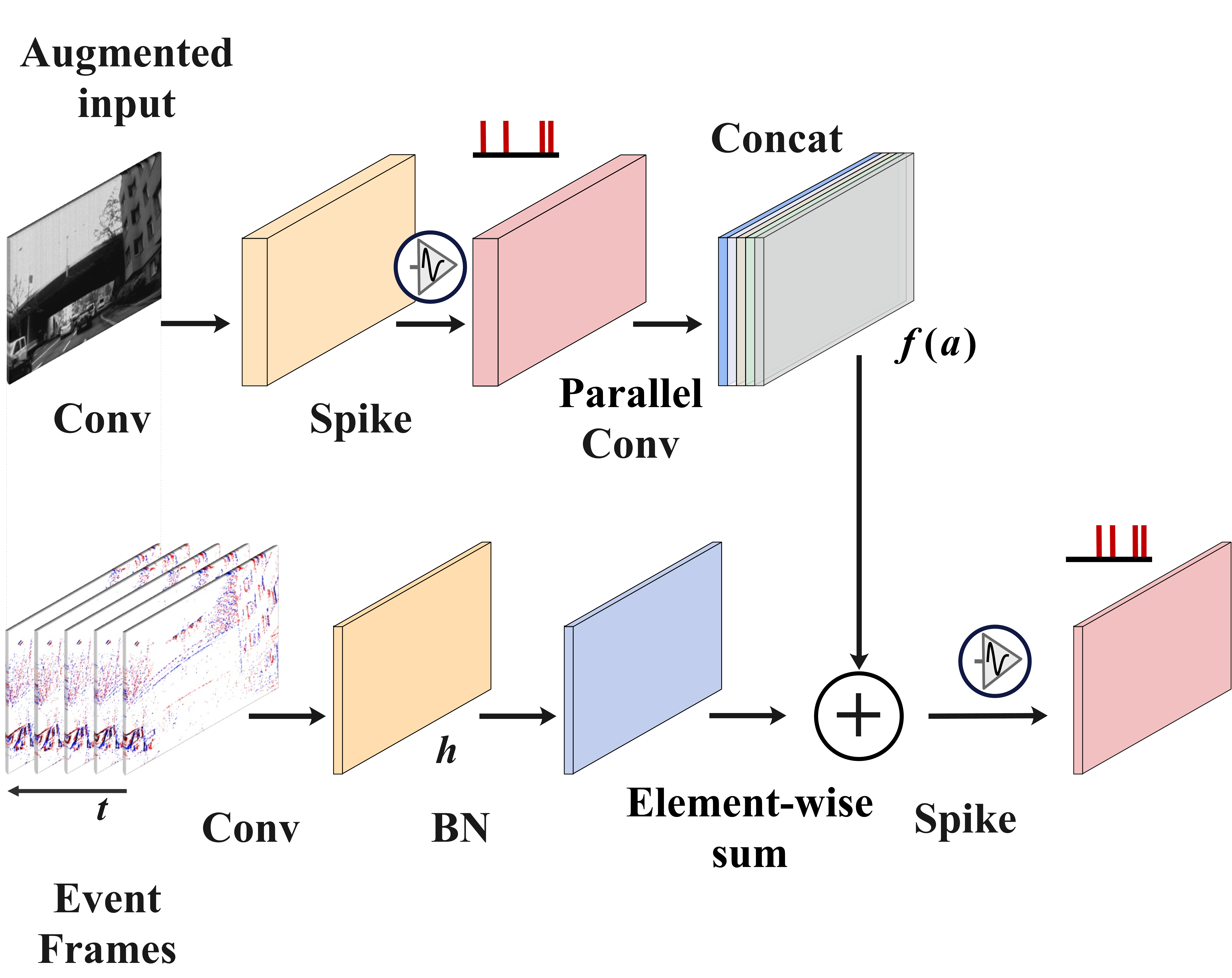}
\caption{Implementation of Spiking Spatially-Adaptive Modulation (SSAM) module. The SSAM module employs a dual-path SNN following MFI to enhance event representation. The augmented input may be original input events, images, or high-quality RGB images. Key components include: \textbf{Conv} (2D convolution operation), \textbf{BN} (batch normalization), \textbf{Spike} (spiking activation), \textbf{Parallel Conv} (multiple parallel dilated convolutions with differing dilation rates), \textbf{Concat} (concatenation operation), and \textbf{Element-wise sum} (addition operation performed between feature maps of the upper and lower paths).}
\label{fig3}
\end{figure}

Prior studies have indicated that a straightforward stack of convolution and normalization operations might dilute semantic information \cite{park2019semantic}. This challenge is mitigated by utilizing pathways with minimal normalization to enrich the network's information flow \cite{cadena2021spade}. 
In order to reduce the loss of semantic information during standard normalization processes, \cite{park2019semantic} proposed the SPADE (Spatially-Adaptive (DE) normalization), which effectively preserves semantic details while transforming input segmentation masks into photorealistic images. 
However, the model’s reliance on computationally expensive multiplication operations poses significant challenges for implementation on neuromorphic hardware, which prefers operations that conform to the MFI principle. 
This limitation underscores the need for a new approach that reduces computational demands while maintaining performance.
To this end, we develop a dual-path SNN with augmented input to enhance event representation. A key advantage of SNNs is their capability for MFI, which reduces computational demand and simplifies hardware implementation. Adhering to this feature, we introduce an MFI-compatible SSAM module, depicted in Fig. \ref{fig3}. Let $\textbf{h}$ be the convolution-extracted feature map from input events, and $\textbf{a}$ the augmented input, which could be either original events or images. The mechanism is defined as:
\begin{align}
\hat{h}_{b,c,y,x} &= \frac{h_{b,c,y,x}-\mu_{c}}{\sigma_c}+f_{c,y,x}(\textbf{a}),\\
\textbf{s} &= S(\hat{\textbf{h}}),
\label{eq:subpixel_crossentropy}
\end{align}
where $b, c, y, x$ represent batch index, channel index, and spatial coordinates respectively; $\mu_c$ and $\sigma_c$ are the mean and standard deviation from batch normalization; $f_{c,y,x}(\textbf{a})$ is the learned parameter modifying the event feature by addition at $(c,y,x)$; $S$ denotes a spiking activation function; and $\textbf{s}$ is the resulting binary feature map after modulation. The modulation parameter generation function uses a multi-layer SNN, as shown in Fig. \ref{fig3}, employing multi-scale dilated convolution for parallel processing and feature map concatenation.
% \textcolor{blue}{
Inspired by the concept in ASPP \cite{DBLP:journals/corr/ChenPSA17}, which uses different atrous rates for effective multi-scale information capture, our parallel convolutions utilize four $3\times3$ convolutions with dilation rates of (1,2,3,4). 
To adhere to the MFI principle, our SSAM module omits the multiplication operation associated with another learned parameter. 
This parameter is generated by a function utilizing a convolution layer. 
By exclusively using addition operations, our model simplifies the computational process while preserving the overall performance of the system. 
In Section~\ref{subsec:ablation}, we compare SSAM with multiplicative SSAM which closely resembles the original design of SPADE. 
The experiment substantiates the efficacy of our approach.

For experiments involving SSAM module, we substitute the first stem layer of the encoder with it. Detailed architecture of the SSAM module and its comparative analysis in different architectural designs are provided by the ablation studies in Section~\ref{subsec:ablation}.
% }

\subsection{Architecture of SpikingEDN}
\begin{figure*}[t]
\centering
\includegraphics[width=0.8\textwidth]{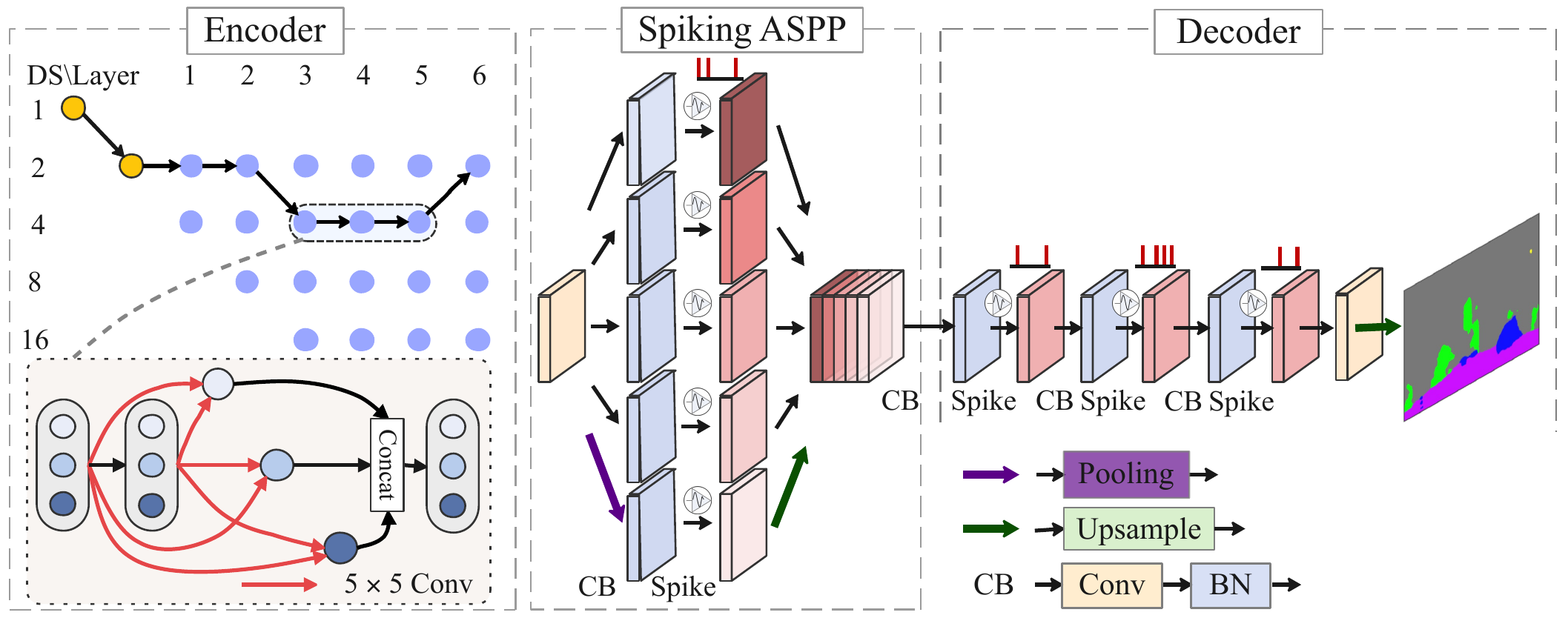}
\caption{
The overall framework of our SpikingEDN. 
\textbf{Top left}: The encoder comprises six layers, involving downsampling and upsampling of the feature map. The black arrow denotes information transmission between layers, involving changes in feature map resolution. Two stem layers, represented by yellow shades, precede the encoder and serve the purpose of channel adaptation and early-stage feature extraction. DS denotes the downsampling rate. \textbf{Bottom left}: The detailed cell structure. The cell forms a directed acyclic graph across three layers, with each layer consisting of three nodes. Each layer receives spike inputs from the previous two layers. Within each layer, the nodes merge inputs from previous layers, and their outputs are concatenated to form the output of the layer. The red arrow represents the layer-to-node operation (5 $\times$ 5 conv). Concat denotes concatenation.
\textbf{Middle}: The spiking Atrous Spatial Pyramid Pooling (ASPP) layer extracts multi-scale features from the encoder and feeds them into the decoder. The spiking ASPP includes four layers each with a 1 $\times$ 1 convolution and three dilated 3 $\times$ 3 convolutions, followed by BNs and spiking activations. The fifth layer features additional pooling before and upsampling after these operations. Outputs from all layers are concatenated and fed into the decoder. 
\textbf{Right}: The decoder comprises a sequence of spiking convolution and BN layers. Finally, an average upsampling layer is employed to refine boundary information and produce the final predicted semantic segmentation map.
}
\label{fig2}
\end{figure*}

In line with standard practices for dense prediction tasks, our approach utilizes a encoder-decoder architecture. Given the critical role of architectural variation in dense estimation networks, special attention is paid to optimizing the encoder, which contains the majority of network parameters and is essential for feature extraction. To this end, we employ a spike-based hierarchical search method \cite{che2022differentiable} to refine the encoder at both cell and layer levels. 

As shown in Fig. \ref{fig2}, two initial spiking stem layers (yellow shades) \cite{che2022differentiable} precede the encoder, serving the purpose of channel adaptation and early-stage feature extraction. The spiking stem layer comprises a convolution operation, a batch normalization (BN) operation, and a spiking neuron treating the former two operations as its input current. 
Notably, the BN operation can be integrated with the convolution during inference \cite{ioffe2015batch}, thus adhering to the MFI principle. 
The searched encoder has a total of six layers, involving downsampling and upsampling of the feature map. The cell forms a directed acyclic graph across three layers, with each layer consisting of three nodes. 
The structure of the cell is repeated across layers, with each layer receiving spike inputs from the previous two layers. 
Within each layer, the nodes are spiking neurons that merge inputs from previous layers, and their outputs are concatenated to form the output of the layer.

During search, the node operation is formulated as:
\begin{equation}
y_j = f(\sum_{i}o^{(i,j)}(y_i)).
\label{eq:node_sum}
\end{equation}
The output spike of node $j$, denoted as $y_j$, is determined by the spiking neuron model $f$, which encompasses the entire equation set of Eq. (2).
$I_j = \sum_{i}o^{(i,j)}(y_i)$ represents the summed inputs from prior nodes or cells, where $y_i$ is the output of a preceding node or cell, and $o^{(i,j)}$ specifies the operation of the directed edge $(i\rightarrow j)$, such as Conv-BN or skip operation. During the search process, each operation $o^{(i,j)}$ is replaced by a weighted average of potential operations $\bar{o}^{(i,j)}(y_i)$:
% }

\begin{equation}
\bar{o}^{(i,j)}(y_i) = \sum_{o \in O^{(i,j)}}\frac{\mathrm{exp}(\alpha_o^{(i,j)})}{\sum_{o \in O^{(i,j)}} \mathrm{exp}(\alpha_o^{(i,j)})}o(y_i),
\label{eq:alpha_op}
\end{equation}
where $O^{(i,j)}$ represents the set of candidate operations for edge $(i\rightarrow j)$, and $\alpha_o^{(i, j)}$ is a trainable continuous variable that acts as the weight for each operation $o$.

The encoder's structure is searched within a predetermined L-layer trellis, where the spatial resolution of each layer can either be halved, doubled, or remain unchanged from the preceding layer. This variation is determined by a set of trainable weighting factors alongside ${\alpha}$. After completing the search process, an optimized architecture is extracted from the trellis. For upsampling operations within the network, nearest interpolation is used to maintain a binary feature map. In ablation study (Section \ref{subsec:ablation}), we evaluate the encoder's efficacy by comparing against the widely used Spiking-ResNet \cite{sengupta2019going, fang2021deep, zheng2021going, hu2021spiking}.

% \textcolor{blue}{
At the end of the encoder, we place an Atrous Spatial Pyramid Pooling (ASPP) layer \cite{DBLP:journals/corr/ChenPSA17} with spiking activations to capture multi-scale features, which is composed of several spiking convolution and BN layers. The first four convolutional layers consist of one 1 $\times$ 1 convolution and three 3 $\times$ 3 convolutions with dilation rates of (6, 12, 18), followed by a BN and spiking activation, respectively.
Unlike the first four layers, the final layer of the spiking ASPP adds a pooling layer before the 1 $\times$ 1 convolution, BN, and spiking activation, and includes an upsampling operation afterwards. 
The features from the five layers of the spiking ASPP will be combined using a concatenate operation and then fed as input to the decoder.

 % An average upsampling layer follows, designed to refine boundary details and construct the final predicted semantic segmentation map.
% }
The decoder architecture comprises three successive spiking convolution layers and a final upsampling layer. This layer plays a crucial role in retrieving boundary information through the learning of low-level features. To achieve smooth classification boundaries in the final segmentation output, the last upsampling layer's output is formulated as floating-point values.
Detailed information on the network structure is provided in the supplementary materials.
% Section \ref{exp_detail}.

\section{Experiments}\label{sec:result}
We evaluate our proposed SNN model for EbSS on the benchmark DDD17 dataset \cite{binas2017ddd17,alonso2019ev} and the recently introduced, high resolution DSEC-Semantic dataset \cite{sun2022ess,gehrig2021dsec}. We employ the prevalent loss function and evaluation metrics in semantic segmentation as in \cite{alonso2019ev,liu2019auto,kim2021beyond}. The loss function is defined as the average per-pixel cross-entropy loss:
\begin{equation}
L=-\frac{1}{N} \sum_{i=1}^{N} \sum_{c=1}^{C} y_{i,c} \ln({\hat{y}_{i,c}}),
\end{equation}
where $N$ is the total number of labeled pixels, and $C$ represents the number of classes. Here, $y_{i,c}$ is the ground truth binary label of pixel $i$ for class $c$, and $\hat{y}_{i,c}$ is the predicted probability by the model. For EbSS, we utilize MIoU as the primary metric. For a given predicted image $\hat{y}$ and a ground truth image $y$, the MIoU is calculated as:
\begin{equation} 
\operatorname{MIoU}(y, \hat{y})=\frac{1}{C} \sum_{c=1}^{C} \frac{\sum_{i=1}^{N} \delta\left(y_{i, c}, 1\right) \delta\left(y_{i, c}, \hat{y}_{i, c}\right)}{\sum_{i=1}^{N} \max \left(1, \delta\left(y_{i, c}, 1\right)+\delta\left(\hat{y}_{i, c}, 1\right)\right)},
\end{equation}
where $\delta$ signifies the Kronecker delta function, and $y_i$ specifies the class of pixel $i$.

% \textcolor{blue}{
This section is organized as follows. We first detail our basic experimental setup and present the main results obtained by the proposed SpikingEDN. We then evaluate the effectiveness of the AiLIF neuron in terms of accuracy, firing rate, and streaming inference. Afterward, ablation studies are conducted to examine the impact of different structures of encoder and SSAM module on the performance of the proposed SpikingEDN. Moreover, to underscore SpikingEDN's suitability for low-power computing, we compare its computational efficiency against other ANNs. We also discuss the outcomes of random seed experiments, which confirm the stability of our final architecture, and demonstrate how incorporating images provides valuable additional information for semantic segmentation.
% Finally, we elaborate on the detailed architecture search and retraining procedures.
The detailed architecture search and retraining procedures are provided in the supplementary materials.
% }

\subsection{Experiment on {DDD17}}\label{subsec:ddd17}
% \ref{subsec:ddd17}
The DDD17 dataset, introduced by \cite{alonso2019ev}, was the first event-based semantic segmentation dataset, derived from the DDD17 driving dataset. It encompasses over 12 hours of recordings from a DAVIS sensor (346$\times$260 pixels) under various driving scenarios, including highways and urban settings. The semantic dataset specifically selects six driving sequences from DDD17 based on criteria like contrast and exposure in generated labels. The training set is composed of five sequences, while the test set includes one sequence. The labels span six categories: flat (road and pavement), background (construction and sky), objects, vegetation, humans, and vehicles.

\subsubsection{Input Representation and Streaming Inference}
In practical scenarios, sensors generate events over various durations continuously. To leverage the SNNs' proficiency in learning from temporal data correlations, we trained the network using a stream of multiple continuous SBT stacks. 
These stacks, together with their successive corresponding labels, formed the continuous target output stream. 
The parameters for each stack were set to $\Delta t = 50\mathrm{ms}$, $n = 5$, and $T = 10\mathrm{ms}$. 
We employed four continuous stacks as a singular input, with the initial stack dedicated to network initialization. 
The output target comprised three temporally consecutive labels, synchronized with the input, enabling the network to generate a label for each simulation step. 
The duration of each SBT input was $50\mathrm{ms}$, incorporating a $10\mathrm{ms}$ merge window during training, thus preserving essential temporal information. This approach allows the SNN to produce one label per step, considerably reducing training time in comparison to previous SNN models \cite{kim2021beyond}. 
Similar strategies have been observed in other applications, such as event-based optical flow estimation, where recent SNN research \cite{hagenaars2021self, schnider2023neuromorphic} has employed analogous settings. 
For evaluation, we utilized the identical input configuration as in training, with an equivalent number of steps for sequential segmentation. 
In Section \ref{subsubsec:ablation}, we investigate the real-time performance and resilience of our SpikingEDN through continuous inference on stacks with varied lengths.

Our AiLIF neuron settings include a $\beta$ value of 0.07 and an initial $\tau_a$ of 0.3, confined to the range [0.2, 0.4] during training. 
This configuration permits a maximal threshold increment of approximately 0.1. In the SSAM module, the AiLIF neuron is applied to the final spiking layer. 

% \begin{figure}
% \centering
% \includegraphics[width=0.5\textwidth]{fig/figure3_searched_0422.pdf} % Reduce the figure size so that it is slightly narrower than the column.
% \caption{Encoder network obtained via searching on DDD17. \textbf{Left}: The 6-layer network architecture, with the black arrow denoting changes in the network layer resolution. \textbf{Right}: The detailed cell structure within the architecture, with each cell comprising three nodes. The red arrows represent operations between each node, as well as between nodes and cells (5 $\times$ 5 conv), while the black arrows indicate data transfer between cells. DS and CC denote downsample rate and concatenate operation, respectively.}
% \label{searched_ddd17}
% \end{figure}

\subsubsection{Architecture Search and Retraining} 
The encoder's architecture search involved a bifurcated approach, which partitions the training dataset into two segments to facilitate a bi-level optimization hierarchical search. 
Within this architecture, each cell comprises three nodes, offering a choice of three operations: skip connection, and convolutions of $3\times3$ and $5\times5$ sizes. 
The defined search space for the layers was capped at six, with a four-level trellis structure incorporating downsampling rates of ${4,2,2,2}$. 
This search process spanned 20 epochs with a batch size of 2. 
The initial 5 epochs were dedicated to initializing the supernet weights, which was then followed by 15 epochs of bi-level optimization. 
The entire search required approximately 2 GPU days. 
The encoder architecture obtained from this search is depicted in Fig. 
\ref{fig2}.
% \ref{searched_ddd17}.

For the retraining phase, the model identified through the search was assigned random initial parameters and underwent training for 100 epochs, including a channel expansion phase. 
This retrained encoder was applied consistently across various input conditions, including pure event inputs as well as augmented scenarios. 
In scenarios involving the SSAM module, both images and identical event stacks with a frame duration of 50 ms were employed as augmented inputs. 
%\textcolor{blue}{Comprehensive details regarding the architecture and the training schedules are further elaborated in the supplementary materials}.
% Section \ref{exp_detail}

\begin{table}[htbp!]
\centering
\caption{Results on the DDD17 Dataset. E denotes events and F denotes gray-scale image for input in test inference. Transfer denotes training with transfer learning. Pre-trained denotes training based on pre-trained network. Direct denotes direct training. Hybrid denotes SNN and ANN hybrid model.} 
%  \resizebox{\linewidth}{!}{
% \resizebox{\textwidth}{15mm}{
 \setlength{\tabcolsep}{1.2mm}{
    \begin{tabular}{lcccrrr}
    \hline
    % \cmidrule(lr){1-3}   %局部中线 横跨2-5列
      Method & Type & Training & Input & \tabincell{c}{Time\\steps}  & \tabincell{c}{Params\\ (M)} & \tabincell{c}{MIoU\\(\%)}\\
    \hline
    % DeepLab(VGG9) & ANN & E & 20 & 4M & 33.7\\
    % FCN(VGG9) & ANN & E & 20 & 13M & 34.2\\
    Evdistill & ANN & Transfer& E  & - & 5.81 & 58.02 \\
    ESS & ANN & Transfer& E & - & 6.69 & 61.37 \\
    CMESS &  ANN &  Transfer &  E &  - &  3.72 &  58.69 \\
    ResNet50+decoder & ANN & Pre-trained & E & - & $>$23 & 59.15 \\
    ESS & ANN & Transfer & E+F & - & 6.69 & 60.43 \\
    CMESS &  ANN &  Transfer &  E+F &  - &  3.72 &  \textbf{64.30} \\
    \hline       
    Ev-SegNet & ANN & Direct& E & - & 29.09 & 54.81 \\
    % ESS & events+frames & - & - & 60.43 \\
    Spiking-DeepLab & SNN & Direct& E & 20 & 4.14 & 33.7\\
    Spiking-FCN & SNN & Direct& E & 20 & 13.60 & 34.2\\
    % Ours (LIF) & SNN & \textcolor{blue}{Direct} & E & 1 & 8.50 & ??? \\
    % \textbf{Ours (Resnet)} & SNN  & E & 1 & 9.1M & \textbf{48.86} \\
    HALSIE &  Hybrid &  Direct &  E+F &  - &  1.82 &  60.66 \\
    Ev-SegNet & ANN & Direct & E+F & - & 29.09 & 68.36 \\
    Ours & SNN  & Direct& E & 1 & 6.74 & 50.23 \\
    Ours & SNN  & Direct& E & 1 & 8.50 & 51.39 \\
    % \textbf{Ours (SSAM)} & SNN  & E & 1 & 8.36M & \textbf{52.08} \\
    Ours (SSAM) & SNN  & Direct& E & 1 & 8.6 & 53.15 \\
    % Evdistill & ANN & A(KD) & - & 5.81M & 72.63 \\
    Ours & SNN  & Direct& E+F & 1 & 8.50 & 61.84 \\
    % \textbf{Ours (SSAM)} & SNN  & E+F & 1 & 8.62M & \textbf{71.27} \\
    Ours (SSAM) & SNN  & Direct& E+F & 1 & 6.41 & 71.68 \\
    Ours (SSAM) & SNN  & Direct& E+F & 1 & 8.62 & \textbf{72.57} \\
    % \textbf{Ours (Resnet+SSAM)} & SNN  & E+F & 1 & 8.27M & \textbf{70.02} \\
    % \textbf{Ours(SSAM)} & SNN  & E+F & 1 & 8.62M & \textbf{72.3} \\
    \hline %底部线
    \end{tabular}}

\label{table1}
\end{table}
\begin{figure*}[t]
\centering
\includegraphics[width=1\textwidth]{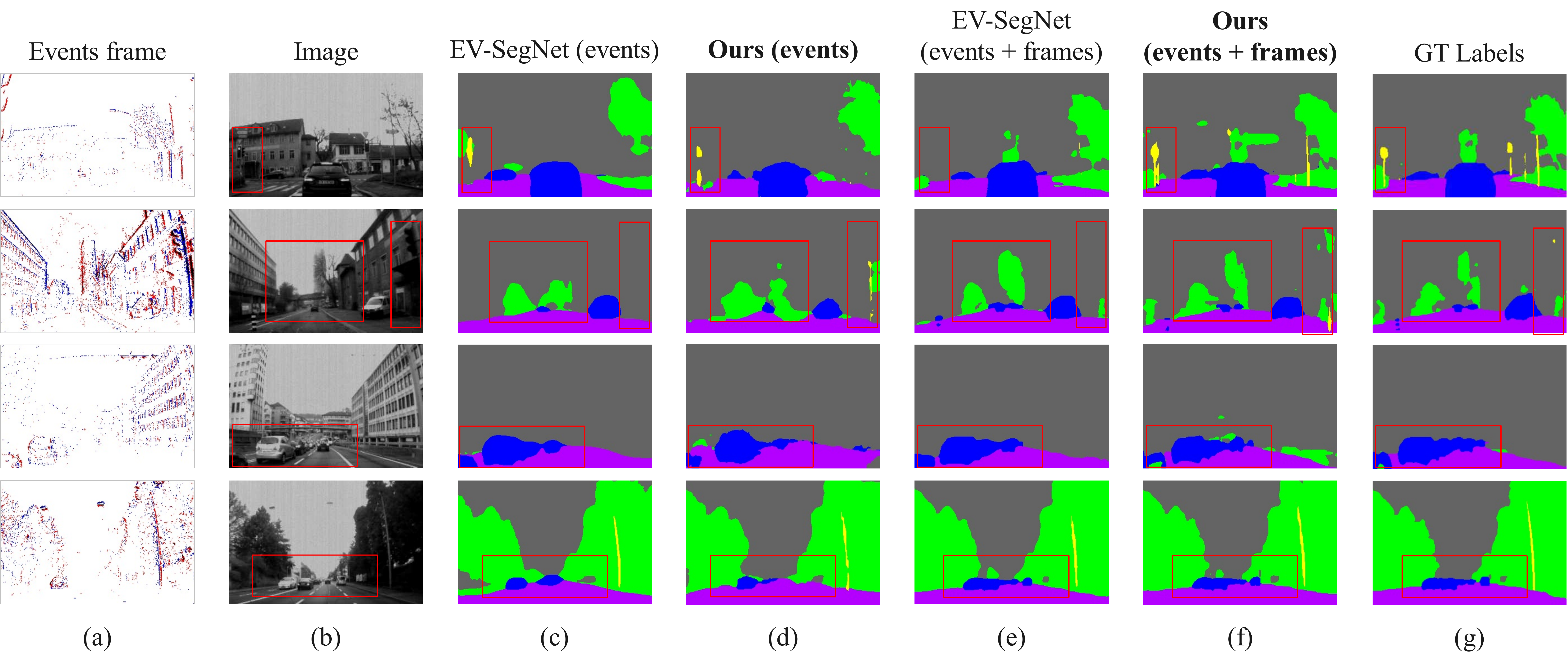}
\caption{Qualitative comparison on the DDD17 dataset. Red boxes in the images highlight areas where our SpikingEDN's predictions align more closely with the ground truth labels. Column (a) visualizes event data processed using the SBT method, while column (b) shows corresponding original grayscale images. Columns (c) and (d) compare the predictions of EV-SegNet and our SpikingEDN using only event data, respectively. Columns (e) and (f) display results from semantic segmentation using both images and events. The final column (g) contains the ground truth labels. Images from EV-SegNet are taken from their paper, whereas our results are based on the SSAM module.}
\label{fig5}
\end{figure*}

\subsubsection{Results}
We benchmarked our SpikingEDN against leading event-based semantic segmentation approaches, encompassing both ANNs and SNNs. 
The comparison included directed trained systems like Spiking-Deeplab/FCN \cite{kim2021beyond}, Ev-SegNet \cite{alonso2019ev}, HALSIE \cite{das2024halsie} and also recent transfer learning approaches including ESS \cite{sun2022ess}, Evdistill \cite{wang2021evdistill}, CMESS \cite{xie2024cross} and contemporary pre-training methods \cite{yang2023event}. Focusing on direct training strategies, our SpikingEDN demonstrates a substantial improvement in MIoU. Specifically, it shows an enhancement of nearly 19\% for pure event inputs and a remarkable 38\% increase for augmented image inputs, surpassing the previous best SNN, Spiking-FCN. This achievement is notable as it is accomplished with a comparatively smaller network and a streamlined inference process requiring only a single time-step, unlike the 20 time-steps demanded by both Spiking-Deeplab and Spiking-FCN. For combined inputs, our SpikingEDN surpasses the current best ANN model, Ev-SegNet, by 4\%, despite that it is three times smaller. While ESS achieves the highest accuracy with pure event inputs using transfer learning, it performs less optimally with combined inputs, 12\% lower than our best result. Even with a similar parameter count as ESS, SpikingEDN significantly outperforms it, demonstrating SpikingEDN's capability to exceed ANN models in MIoU.
HALSIE's SNN and ANN hybrid approach achieves a 60.66\% mIoU on the DDD17 dataset, trailing our results by 11.91\%. Meanwhile, CMESS, employing a transfer learning approach, reaches a peak mIoU of 64.30\%, still 8.27\% lower than our performance. 
Note that based on knowledge distillation and lightweight architectures, both CMESS and HALSIE have significantly fewer parameters. Similar techniques could also be applied to SNNs to further improve model efficiency, as demonstrated in a recent work \cite{shen2023esl}.
A qualitative comparison of semantic segmentation images is shown in Fig. \ref{fig5}, with red boxes indicating areas where SpikingEDN's predictions more closely match the ground truth labels. For instance, the traffic sign in the top row of Fig. \ref{fig5} is more accurately recognized by SpikingEDN compared to Ev-SegNet.
Moreover, implementing SSAM module in place of the first stem layer significantly enhanced network performance for both event-only and grayscale-augmented inputs, particularly, improving the accuracy by nearly 2\% with event-only inputs.

\begin{figure*}
\centering
\includegraphics[width=1\textwidth]{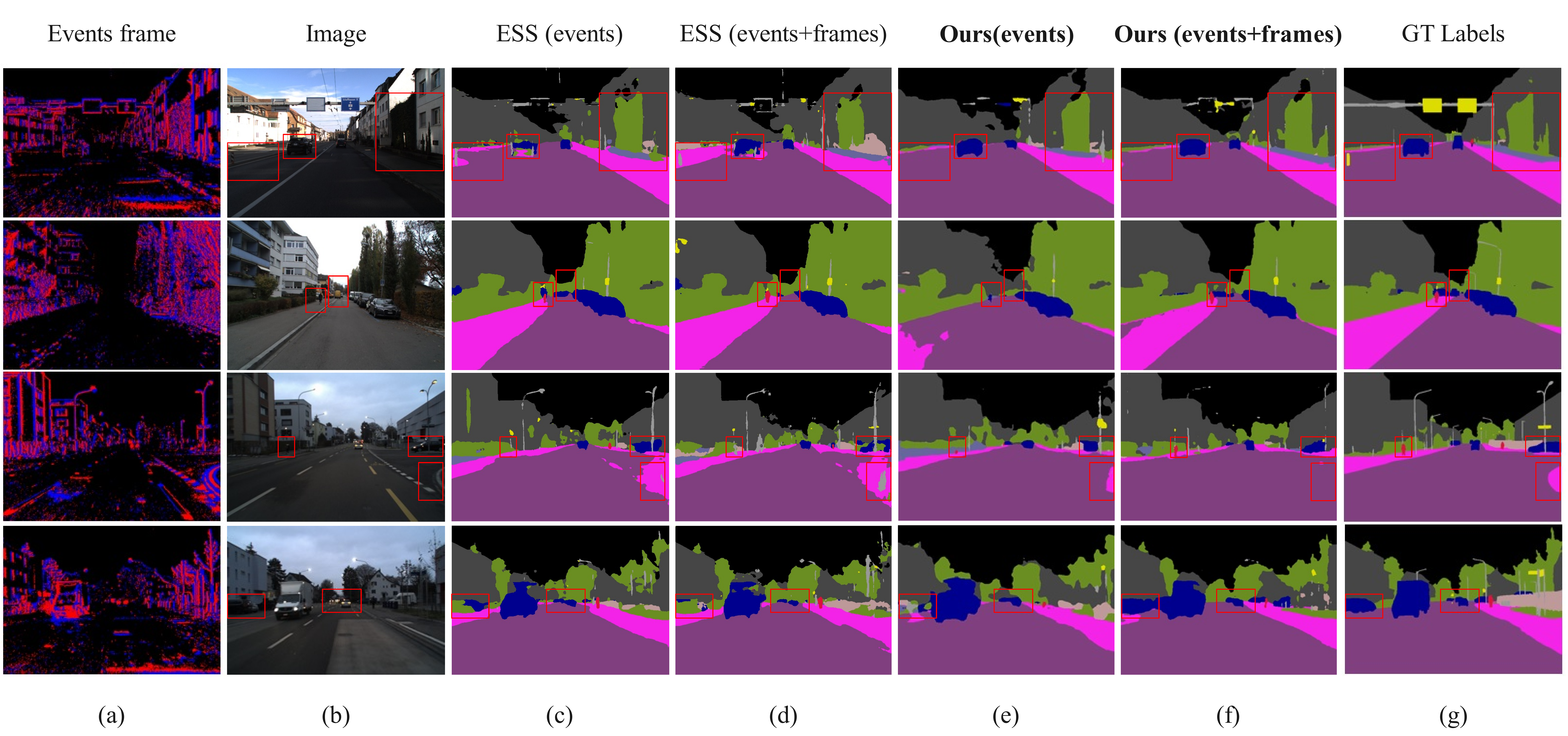} 
\caption{Qualitative comparison on the DSEC-Semantic dataset. Our SpikingEDN, which combines events and frames as input, accurately captures certain details (marked area) better than other methods. The red boxes in the images denote areas where SpikingEDN's predictions closely match the ground truth labels. Figures in columns (a) and (b) represent visualizations of event data and RGB images, respectively. Diagrams in columns (c) and (d) depict the results of the ESS method with purely event-based input and combined event and image inputs, respectively. Columns (e) and (f) compare the predictions of SpikingEDN using solely event data and a combination of RGB images and events (with the SSAM module), respectively. The last column (g) contains the ground truth labels.}
% \textcolor{blue}{ESS with events+frames?}
\label{fig_dsec}
\end{figure*}

\subsection{Experiment on DSEC-Semantic}

% more example had been added  20230524
\begin{table}[htbp!]
\centering
\caption{Results on the DSEC Dataset. E denotes events and F denotes RGB image for input in test inference. Transfer denotes training with transfer learning. Pre-trained denotes training based on pre-trained network. Direct denotes direct training. Hybrid denotes SNN and ANN hybrid model.}
 \setlength{\tabcolsep}{1.2mm}{
    \begin{tabular}{lrccrrr}
    \hline 
    Method & Type & Training & Input & \tabincell{c}{Time\\steps}  & \tabincell{c}{Params\\ (M)} & \tabincell{c}{MIoU\\(\%)}\\
    \hline
    ESS & ANN & Transfer & E & - & 6.69 & 51.57 \\
    CMESS &  ANN &  Transfer &  E &  - &  3.72 &  57.49 \\
    ResNet50+decoder & ANN & Pre-trained & E & - & $>$23 & 59.16 \\
    ESS & ANN & Transfer & E+F & - & 6.69 & 53.29 \\
    CMESS &  ANN &  Transfer &  E+F &  - &  3.72 &  \textbf{59.53} \\
    \hline %中部线
    Ev-SegNet & ANN & Direct &$\text{E}$ & - & 29.09 & 51.76 \\
    HALSIE &  Hybrid &  Direct &  E+F &  - &  1.82 &  52.43 \\
    Ours & SNN  & Direct & E & 1 & 6.27 & 53.17 \\
    Ours (LIF) & SNN & Direct & E & 1 & 8.50 & 52.71 \\
    Ours & SNN  & Direct & E & 1 & 8.50 & 53.04 \\
    % \hline %中部线
    Ours (SSAM) & SNN  & Direct & E+F & 1 & 6.42 & 57.73 \\
    Ours (SSAM+LIF) & SNN  & Direct & E+F & 1 & 8.63 & 57.22 \\
    Ours (SSAM) & SNN  & Direct & E+F & 1 & 8.63 & 57.77 \\
    Ours (SSAM) & SNN  & Direct & E+F & 1 & 23.09 & \textbf{58.32} \\
    \hline 
    \end{tabular}}

\label{table2}
\end{table}
The DSEC-Semantic dataset \cite{sun2022ess}, an extension of the DSEC dataset \cite{gehrig2021dsec}, features diverse recordings from both urban and rural settings. It provides labels of 640$\times$440 pixels resolution across 11 classes, delivering higher quality and more intricate labeling compared to the DDD17 dataset. Our approach to partitioning the training and test sets aligns with the methodology outlined in \cite{sun2022ess}. % \textcolor{blue}{
In this part, we employed the same network architecture as on the DDD17 dataset and trained them directly on the DSEC-Semantic dataset. 
To maintain parity with recent advancements in the field \cite{yang2023event}, we expanded the channel dimensions of our SpikingEDN. 
Across all input configurations, SpikingEDN demonstrated superior performance over the ESS model \cite{sun2022ess}, which relies on transfer learning, particularly when comparing models with similar parameter counts. 
Notably, when utilizing combined event and frame inputs, SpikingEDN achieved an MIoU of 57.73\%, exceeding the ESS model by over 4\%.
Our approach also outperforms HALSIE \cite{das2024halsie} based on direct training, showing an improvement of 5.89\% in mIoU. 
The recent work proposed in \cite{yang2023event} marginally surpasses our model by less than 1\% with a comparable parameter count. It leverages a pre-trained ResNet50 ANN encoder, which was trained through contrastive learning and augmented with the ImageNet-1K RGB dataset. 
The very recent CMESS model \cite{xie2024cross}, which utilizes an ANN-based knowledge distillation approach, surpasses our results by 1.21\%.
These results underscore SpikingEDN's scalability and proficiency in handling high-resolution event-based vision tasks, competitive to sophisticated ANNs. 
Fig. \ref{fig_dsec} provides a qualitative analysis, showcasing SpikingEDN's enhanced detail capture in specific scenarios, such as the person and car highlighted in the second and final rows, respectively. 
% }

\subsection{AiLIF Neuron and Streaming Inference} 
\label{subsubsec:ablation}
\begin{figure*}[!htbp]
\centering
\includegraphics[width=1\textwidth]{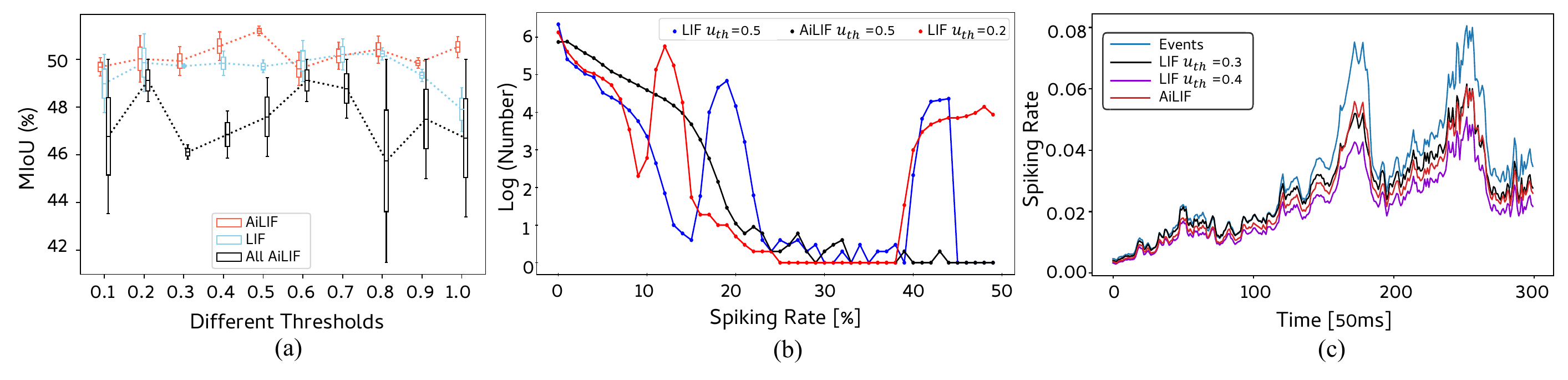}
\caption{(a) Comparisons of network performances on the DDD17 dataset by applying LIF, AiLIF neuron on the first layer, or AiLIF neuron to the whole network, on a range of threshold values. Network training takes two different random seeds. (b) Distribution of spiking rates in the first layer. The horizontal axis represents the spiking rate, while the vertical axis denotes the number of neurons in a specific firing rate range. The vertical axis utilizes a logarithmic scale with a base of 10. (c) Illustration of event density and activation of AiLIF and LIF neurons in the first layer during inference for a short time. The activation of the AiLIF neuron (base threshold 0.3) is generally between the two LIF neurons of boundary thresholds. Note that its activation surpasses the LIF neuron with a 0.3 threshold in the first peak of event density, revealing the intricacy of dynamic modulation.}
\label{fig_alif_distribution}
\end{figure*}

To evaluate the influence of adaptive threshold in event processing, we compared AiLIF with standard iterative LIF neurons.
Results on the DSEC-Semantic dataset show that networks incorporating AiLIF neurons in either the first or the final layer of SSAM module achieved increase in accuracy by 0.33\% and 0.55\%, respectively,  as opposed to those using only LIF neurons. 
We extended this evaluation to various threshold settings on the DDD17 dataset, specifically under pure event input conditions. As depicted in Fig. \ref{fig_alif_distribution}a, the results reveal that networks with AiLIF neurons in the first layer consistently maintained stable accuracy across different threshold values, typically outperforming networks with exclusive LIF neuron usage.
However, networks that employed AiLIF neurons in all layers exhibited a marked decrease in performance. 
This observations corresponds with findings from \cite{hagenaars2021self}, where the deployment of ALIF neurons throughout an entire network resulted in less optimal outcomes compared to the utilization of purely LIF neurons. 
These results imply that excessive flexibility in neuron configuration might adversely affect the precision of training, thus meriting further exploration in future research.

Additionally, the self-adaptive threshold feature of AiLIF neurons not only improved accuracy but also contributed to a more balanced and generally lower spiking rate in the first layer, as illustrated in Fig. \ref{fig_alif_distribution}b. We compared AiLIF neurons against a standard LIF neuron at a baseline threshold of 0.5 and a LIF neuron adjusted for comparable accuracy (achieving 51.07\% at a threshold of 0.2). The AiLIF neuron effectively regulated and moderated information flow into subsequent network layers, resulting in reduced firing rates and, consequently, lower energy consumption across the network, as detailed in Section IV-E. This observation suggests that when strategically integrated, AiLIF neurons can enhance both the accuracy and computational efficiency of SNNs. This insight is particularly significant in light of previous research in event-based dense prediction, which indicated that networks incorporating ALIF neurons in all layers demonstrated subpar results \cite{hagenaars2021self}.

Fig. \ref{fig_alif_distribution}c provides a visual representation of input event streams according to their density, alongside the activation responses of various neurons. Notably, the LIF neuron with a threshold of 0.4, roughly equivalent to the upper limit of the AiLIF neuron's adaptive threshold, exhibits the lowest activation. The AiLIF neuron (with a base threshold of 0.3) generally shows activation levels between two LIF neurons with fixed thresholds, highlighting the nuanced nature of its dynamic response. Interestingly, during periods of high event density at the onset, the AiLIF neuron's activation surpasses that of the LIF neuron set at the same threshold, indicating the complexity and effectiveness of its adaptive modulation.

% \begin{figure}[!htbp]
% \centering
% \includegraphics[width=0.49\textwidth]{fig/0612_density.pdf}
% \caption{Distribution of spiking rates of LIF neurons (the blue bar) and AiLIF neurons (the yellow bar)in the first stem layer. {The horizontal axis represents the spiking rate, while the vertical axis denotes the number of neurons in a specific discharge rate range. Moreover, the vertical axis utilizes a logarithmic scale with a base of 10.}}
% \label{fig_alif_distribution}
% \end{figure}
% \begin{figure}[t]
% \centering
% \includegraphics[width=0.35\textwidth]{fig/Figure_6_inference_0103.pdf} % Reduce the figure size so that it is slightly narrower than the column.
% \caption{Illustration of event density and activation of AiLIF and LIF neurons in the first stem layer during inference for a short time. The activation of the AiLIF neuron (base threshold 0.3) is generally between the two LIF neurons of boundary thresholds. Note that its activation surpasses the LIF neuron with a 0.3 threshold in the first peak of event density, revealing the intricacy of dynamic modulation. }
% \label{sparsity_time}
% \end{figure}
% comparsion of results (alif and lif ) on ddd17 and dsec
%\noindent \textbf{Streaming inference}
\begin{figure}[!htbp]
\centering
\includegraphics[width=0.49\textwidth]{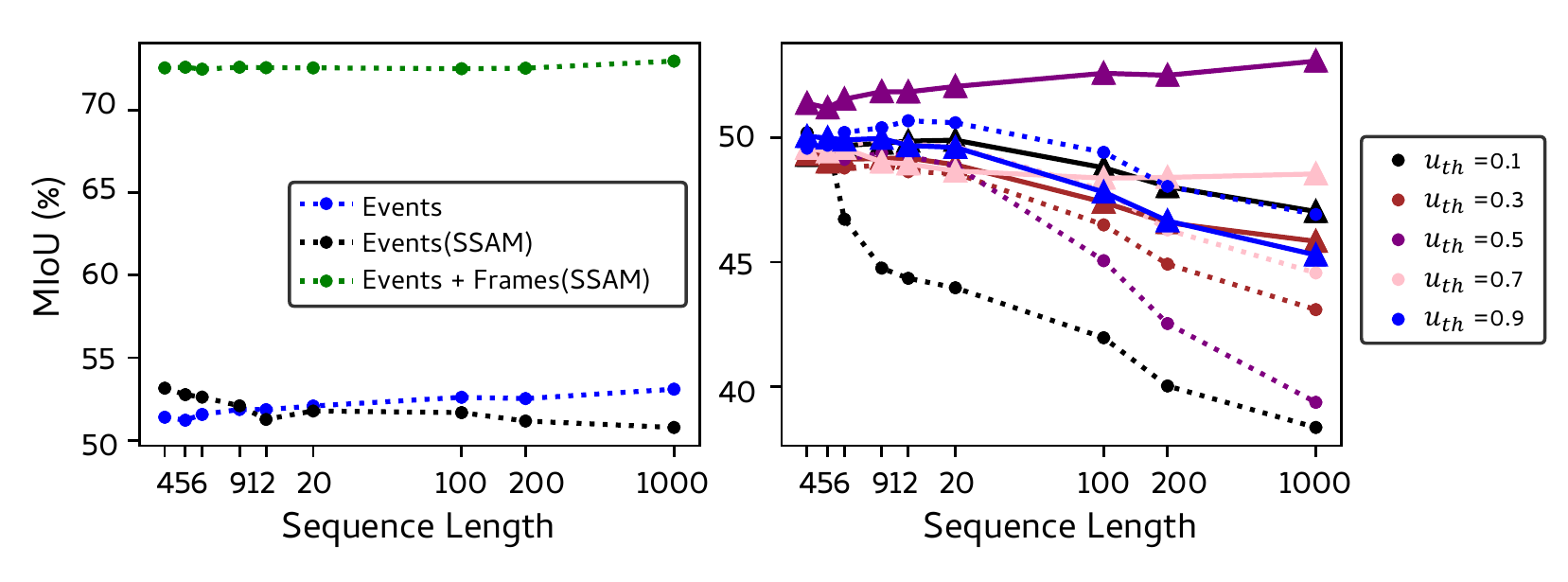} 
% Reduce the figure size so that it is slightly narrower than the column.
\caption{Results of streaming inference on the DDD17 dataset. The sequence length in x axis denotes the continuous inference time steps of the network. Left: SpikingEDN with AiLIF neuron (with threshold 0.5) on the first layer under different input configurations, with or without SSAM module. Right: Comparison of using AiLIF (triangles with solid lines) and LIF neuron (circles with dashed lines) on the first layer of the network under pure events input without SSAM module, across different threshold values. 'Frames' denotes input with images.}
\label{inference_ddd17_dsec}
\end{figure}
To further evaluate the robustness of our SpikingEDN in scenarios featuring real-time variable event stream lengths (i.e., streaming inference), we conducted an assessment on the DDD17 dataset. 
Specifically, we amalgamated the public test set into a singular continuous event stream, encompassing 3890 frames, with each frame representing a SBT stack. 
We then subjected SpikingEDN (with either pure event inputs, with or without SSAM module, or combined event and image inputs using SSAM module) to continuous inference over $T$ steps on this data stream, iterating until the end of the stream. 
As shown in Fig. \ref{inference_ddd17_dsec}, our models demonstrate consistent performances across diverse inference durations, thereby validating their effectiveness and resilience in real-time applications. 
Further, a comparative analysis between networks incorporating AiLIF and standard LIF neurons in the first layer, under various threshold settings, revealed that AiLIF neurons yield more uniform performance over different input stream lengths. 
This finding highlights an additional merit of employing the adaptive threshold methodology in such applications.

% add reason ??? 
\subsection{Ablation Studies}\label{subsec:ablation}
To validate the key design choices of our SpikingEDN and to understand the impact of individual components on overall performance, we carried out comprehensive ablation studies. 
First, we evaluate SpikingEDN's custom encoder against other prevalent spiking encoder architectures.
Second, we conduct a comparative analysis with the architecture searched end-to-end by SpikeDHS. Third, we compare the performance of the MFI-compatible SSAM with the multiplicative SSAM.
Finally, we delve into the architecture of SSAM module, assessing its effectiveness in enhancing event representation for downstream tasks and examining the influence of various configurations of its upper path on network performance. 

\subsubsection{Spiking-ResNet as Encoder} 
\begin{table}[htbp!]
\centering
\caption{Results of Spiking-Resnet as encoder on the DDD17 and DSEC-Semantic Dataset . E denotes events, F denotes images.}
 \setlength{\tabcolsep}{1.2mm}{
    \begin{tabular}{lrcrr}
    \hline 
    Method & Dataset & Input  & \tabincell{c}{Params\\ (M)} & \tabincell{c}{MIoU\\(\%)}\\
    \hline
    Spiking-ResNet & DDD17& E  & 9.10 & 48.86 \\
    Ours & DDD17 & E  & 8.50 & \textbf{48.94} \\
    \hline
    % Spiking-ResNet& S2M & DDD17& E  & 13.05 & 48.11 \\
    % Spiking-ResNet& S2M(nd) & DDD17& E & 9.25 & 47.09 \\
    Spiking-ResNet (SSAM)  & DDD17  & E+F  & 8.27 & 70.02 \\
    Ours (SSAM) &DDD17 & E+F & 8.62 & \textbf{72.30} \\   
    % Spiking-ResNet (SSAM)& S2M& DDD17  & E+F &  8.73 & 69.42 \\
    % Spiking-ResNet (SSAM)& S2M(nd)& DDD17  & E+F  &8.85  & 64.43 \\
    \hline %中部线
     % Ours & SNN  & \textcolor{blue}{Direct} & E & 1 & 8.50 & 53.04 \\
    Spiking-ResNet & DSEC  & E & 8.74 & 50.67 \\
    Ours & DSEC & E & 8.50 & \textbf{52.71} \\
    \hline %中部线
    % Spiking-ResNet& S2M& DSEC  & E & 8.75 & 49.34 \\
    % Spiking-ResNet& S2M(nd)& DSEC  & E & 8.68 & 48.56 \\
    Spiking-ResNet (SSAM) & DSEC  & E+F  & 8.43 & 54.75 \\
    % Spiking-ResNet (SSAM)& S2M& DSEC  & E+F  & 8.74 & 55.29 \\
    % Spiking-ResNet (SSAM)& S2M(nd)& DSEC  & E+F  & 8.36 & 52.60 \\
    Ours (SSAM) & DSEC & E+F & 8.63 & \textbf{57.22} \\
    % Ours (SSAM) & SNN  & \textcolor{blue}{Direct} & E+F & 1 & 8.63 & 57.77 \\
 
    \hline 
    \end{tabular}}
\label{table_resnet}
\end{table}

Derived from the original ResNet \cite{he2016deep}, the Spiking-ResNet has gained popularity for classification tasks, showing performances competitive with ANNs on benchmark datasets \cite{sengupta2019going, fang2021deep, zheng2021going, hu2021spiking}. We compared this encoder with our custom-searched encoder network, ensuring similar parameter counts and following analogous layer variation schemes. To ensure a fair comparison, all networks used LIF neurons, and M2M residual connections \cite{vicente2022keys} were applied to the Spiking-ResNet for alignment with SpikingEDN. Table \ref{table_resnet} presents the results of this comparison. It reveals that on both the DDD17 and DSEC datasets, and across various input configurations, with or without SSAM module, SpikingEDN consistently surpasses the traditional Spiking-ResNet. Notably, the implementation of SSAM module also significantly improves network performance under a different encoder architecture.

% nd means without down sampling
\subsubsection{Comparison with SpikeDHS} 
To demonstrate the effectiveness of our model, we conduct a comparative analysis with SpikeDHS under the EbSS task on the DDD17 dataset. 
To ensure a fair comparison, the SpikeDHS model employs the same encoder structure with the SpikingEDN model, with all spiking neurons configured as LIF neurons. 
For the SpikeDHS network, we substitute the Spiking ASPP and the decoder of the SpikingEDN with 1 $\times$ 1 or 3 $\times$ 3 convolutions, followed by an upsampling operation. 
Additionally, all training setups are kept identical to those used for the SpikingEDN to ensure consistency across experiments. 
This modification allowed us to isolate and evaluate the contribution of the advanced components in SpikingEDN.

The performance is assessed on two input conditions: event-only and grayscale-augmented inputs, excluding SSAM. 
Under these settings, SpikeDHS achieved mIoU scores of 32.81 \% and 51.88 \% with the 1 $\times$ 1 convolution, and 33.25 \% and 52.45 \% with the 3 $\times$ 3 convolution, respectively. 
These results are significantly lower than those achieved by SpikingEDN, which outperforms SpikeDHS by margins of 18.58 \% and 20.69 \% with 1 $\times$ 1 convolution, and 18.14 \% and 20.12 \% with 3 $\times$ 3 convolution, respectively. 
These findings conclusively demonstrate that the enhancements incorporated into SpikingEDN provide substantial improvements over the basic SpikeDHS model, validating the effectiveness of our proposed architecture.

\subsubsection{Comparison with multiplicative SSAM} 
To demonstrate the effectiveness of SSAM, we design experiments comparing SSAM and multiplicative SSAM. 
The multiplicative SSAM is defined as:
\begin{align}
\hat{h}_{b,c,y,x} &= \gamma\frac{h_{b,c,y,x}-\mu_{c}}{\sigma_c}+f_{c,y,x}(\textbf{a}).
% ,\\
% \textbf{s} &= S(\hat{\textbf{h}}),
\label{eq:subpixel_crossentropy_multi}
\end{align}
Closely resembling the original design of SPADE, the multiplicative SSAM introduces an additional learned parameter, $\gamma$, generated by a convolution operation. 
This parameter is applied post the convolution and spike activation in the upstream pathway and is multiplied with the features processed by the downstream pathway and then adds to the other learned parameter $f_{c,y,x}(\textbf{a})$. 
Here, $b, c, y, x$ denote batch index, channel index, and spatial coordinates, respectively, while $\mu_c$ and $\sigma_c$ represent the mean and standard deviation from batch normalization. 
The function $f_{c,y,x}(\textbf{a})$ refers to the learned parameter that modifies the event feature by addition at $(c,y,x)$.

The experiments on the DDD17 dataset compare the network performance with event-augmented and grayscale-augmented inputs. 
The multiplicative SSAM achieved mIoU scores of 51.73 \% and 72.60 \% for event-augmented and grayscale-augmented inputs, respectively. 
In contrast, our MFI-SSAM, without complex multiplication operations, achieved an mIoU of 72.57\% for grayscale-augmented inputs, nearly matching the multiplicative SSAM, and surpassed it with event-augmented inputs at 53.15\%.
These results demonstrate that our MFI-SSAM not only reduces the model size and saves energy by eliminating complex multiplication operations but also meets or even surpasses the performance of the multiplicative SSAM. 
Our approach lifts the restriction of multiplication, thereby allowing SSAM to be directly deployed on potential neuromorphic chips that favors addition operations for faster processing speeds.

\subsubsection{Architecture of SSAM module} 
The dual-path configuration of the SSAM module can replace the first stem layer of the encoder, thereby augmenting event representation for subsequent tasks. The lower path of the module processes 5-channel event frames, while the upper path handles 1-channel augmented data (either an image or the same event stack with a 50 ms frame duration). The output is a 64-channel binary feature map, maintaining the same spatial resolution as the input. The rest of the network's architecture remains as previously described. Table \ref{tab:SSAM_details} details the SSAM module's structure.

\begin{table*}[htbp!]
\centering
\caption{Detailed operations of SSAM module. BN denotes batch normalization. The feature maps from the two paths are fused by addition following a spike activation.}
    \begin{tabular}{clc}\hline
    \textbf{Path}&\textbf{Layer Description}&\textbf{Feature map size}\\\hline
    \multirow{3}{*}{\tabincell{c}{\textbf{Upper}  }}
    & 2D conv. $1\times3\times3\times64$ with BN &$ 64\times200\times346$ \\
    & Spike activation &$ 64\times200\times346$ \\
    & Parallel conv. ($4\times2D$ conv. $64\times3\times3\times$ 16 with dilation \{1, 2, 3, 4\} and concatenation) &$ 64\times200\times346$ \\
    \hline
    \multirow{1}{*}{\tabincell{c}{\textbf{Lower}  }}
    & 2D conv. $5\times1\times1\times64$ with BN &$ 64\times200\times346$ \\
    \hline 
    \end{tabular}
\label{tab:SSAM_details}
\end{table*}

\begin{figure*}[htbp!]
\centering
\subfloat[Upper path structure S1 of SSAM module.] {
 \label{fig_ssam_conv1} 
\includegraphics[width=0.45\textwidth]{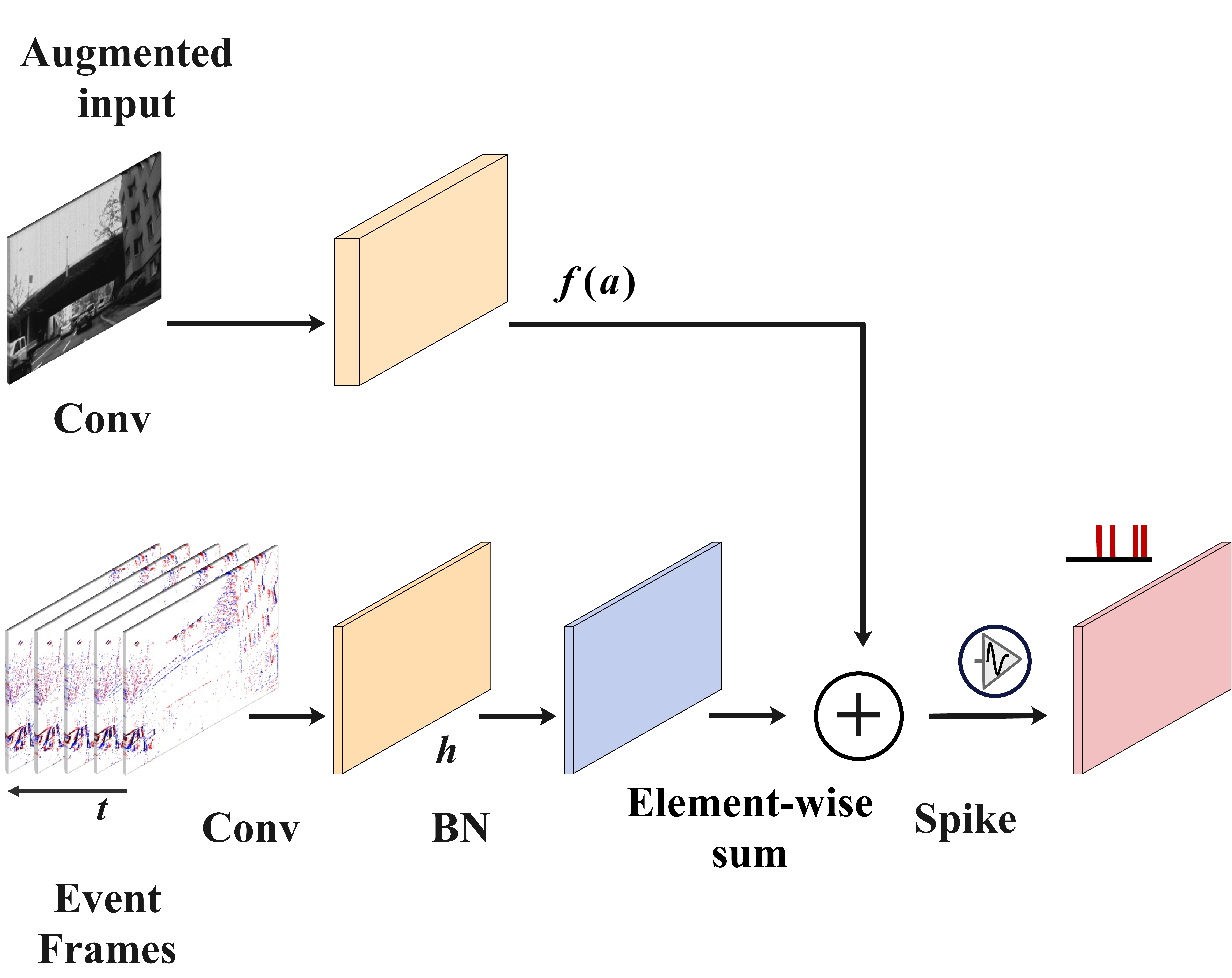} 
 \hspace{3mm}
}
\subfloat[Upper path structure S2 of SSAM module.] { 
\label{fig_ssam2}     
\includegraphics[width=0.45\textwidth]{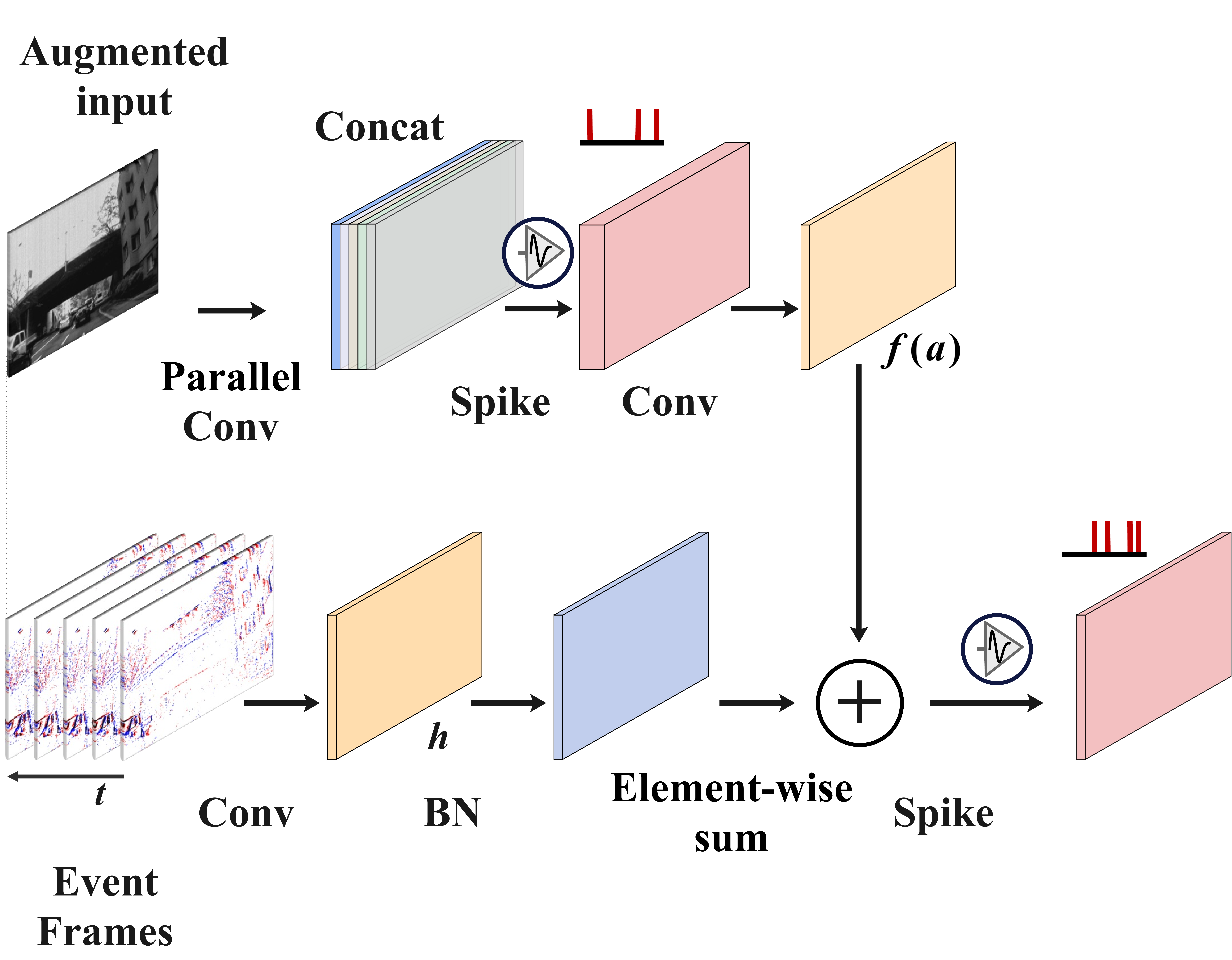}    
}
\caption{
Illustration of the upper path structures S1 and S2 of SSAM module. The upper path can accommodate various inputs, such as original event data, images, or high-quality RGB images. Key components include: \textbf{Parallel Conv} (several concurrent dilated convolutions with differing dilation rates), \textbf{Conv} ( 2D convolution operation), \textbf{BN} (batch normalization), `\textbf{Spike} ( spiking activation), \textbf{Concat} (concatenation operation), and \textbf{Element-wise sum} (addition operation conducted between the feature maps of the upper and lower paths).
Specifically, structure S1 consists of a single convolution layer, while structure S2 comprises a sequence of parallel convolution, spiking activation, and convolution operations.
}     
\label{fig}     
\end{figure*}

Moreover, we assess the network performance with different upper path designs in SSAM module under combined input conditions. These variations include S1: a single convolution layer; S2: parallel convolution followed by spiking activation and another convolution; S3: a convolution followed by spiking activation and then parallel convolution. The different architectures of SSAM module, specifically upper path structures S1 and S2, are illustrated in Figs. \ref{fig_ssam_conv1} and \ref{fig_ssam2}, respectively.
Table \ref{tab:ssam_compare} summarizes the results, indicating that our design, namely S3, outperforms the other configurations. Interestingly, adding a batch normalization layer after the convolution in S1 resulted in a 1\% decline in network performance. This observation validates the efficacy of the design principles underlying the SSAM module.

\begin{table}[!htbp]
\centering
\caption{Results comparison on the DDD17 dataset with different architectures of SSAM module.}
\begin{tabular}{c|ccc}
\toprule 
Upper path structure & S1 & S2 & S3 \\ 
\midrule
MIoU \% & 67.54 & 68.49 & 72.57 \\
\bottomrule
\end{tabular}
\label{tab:ssam_compare}
\end{table}
% \begin{table}[htbp!]
% \centering
% \caption{Results of streaming inference on the DDD17 and Dsec-Semantic Dataset. E denotes events, \textcolor{blue}{F} denotes gray-scale image.}
%  \setlength{\tabcolsep}{1.2mm}{
%     \begin{tabular}{lrcrrr}
%     \hline 
%     Method & Dataset & Input & \tabincell{c}{Time\\steps}  & \tabincell{c}{Params\\ (M)} & \tabincell{c}{MIoU\\(\%)}\\
%     \hline
%     Ours & DDD17& E & 1 &  & \textbf{} \\
%     Ours (SSAM) & DDD17  & E+F & 1 &  & \textbf{} \\
%     \hline %中部线
%     Ours & Dsec  & E& 1 &  & \textbf{} \\
%     Ours (SSAM) & Dsec  & E+F & 1 &  & \textbf{} \\
%     \hline 
%     \end{tabular}}

% \label{table_inference}
% \end{table}

\subsection{Sparsity and Computational Cost}
\begin{figure}[t]
\centering
\includegraphics[width=0.49\textwidth]{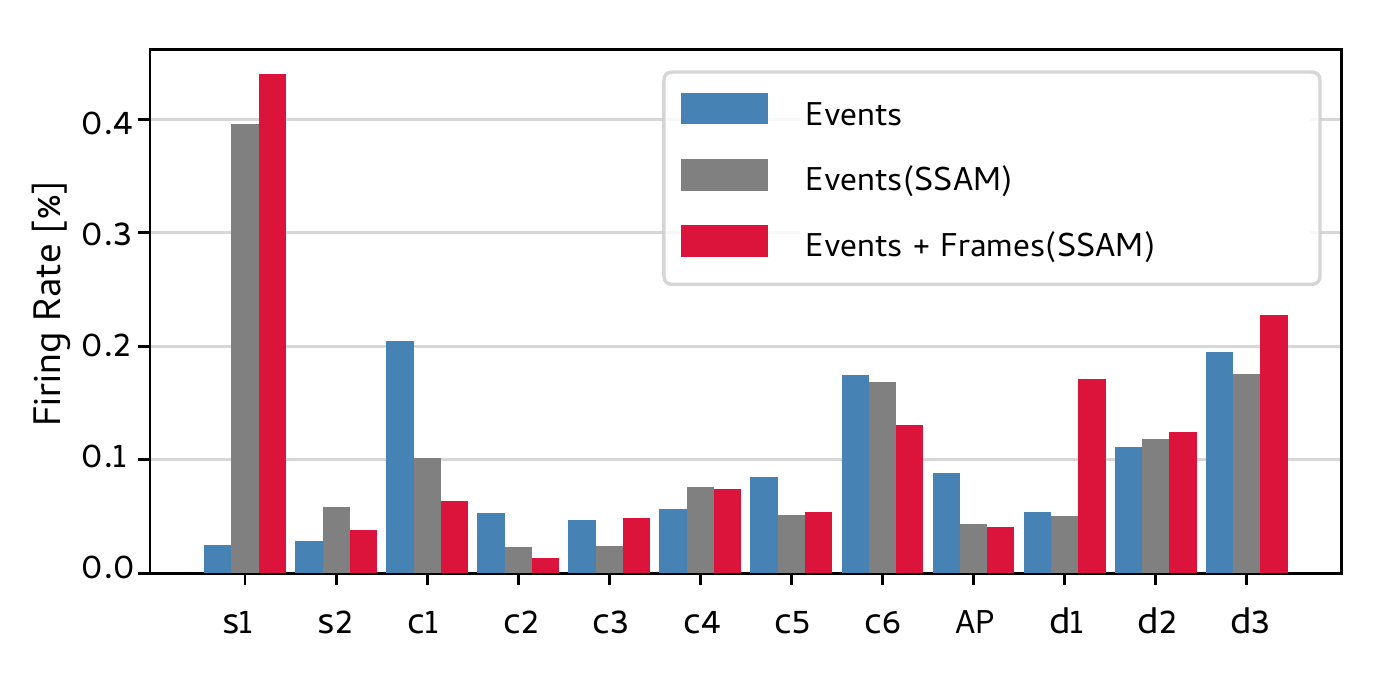}
\caption{Layer-wise sparsity of SNNs on the DDD17 dataset. The horizontal axis represents the different layers of the network, e.g. with 's1','c1', 'AP' and 'd1' denoting the first stem layer, the first cell, ASPP layer and the first decoder layer. The vertical axis depicts the mean firing rate of the layer. `Events' denotes inference with pure events input. `Frames' represents input with images.}
\label{sparsity_hist}
\end{figure}
\begin{table}[!htbp]
\centering
\caption{The operation number and energy cost of different networks on the DDD17 dataset. E and F denote events and gray-scale images, respectively, for input in inference. \#Add. and \#Mult. denote the number of addition and multiplication operations, respectively. FR denotes the mean firing rate of the model.}
\label{tab:energy}
 \setlength{\tabcolsep}{1.2mm}{
\begin{tabular}{lrrrrc}
\toprule Method& Input & {\#Add.} &{\#Mult.} & {Energy} & {FR}\\ 
\midrule
EV-SegNet (ANN)  &E  & 9322M &9322M & 42.88 $\mathrm{mJ}$ & -\\
ESS (ANN)  &E  & 11700M &11700M & 53.82 $\mathrm{mJ}$& -\\
EvDistill (ANN)  &E  & 29730M &29730M & 136.76 $\mathrm{mJ}$& -\\
\midrule
Ours (LIF $u_{\mathrm{th}=0.2}$) & E & 7743M & 17M & 7.03 $\mathrm{mJ}$ &0.102  \\   % stem0111 lif vth0.2 
Ours (LIF) & E &6908M &17M& 6.28 $\mathrm{mJ}$ & 0.091 \\   %vth 0.5 
Ours (AiLIF) & E & 5921M & 17M &	5.39 $\mathrm{mJ}$ & 0.078 \\   % stem0111  vth0.5
% Ours (LIF) E 5766M 17M 5.27 mJ 0.077    vth 0.5
% Ours (AiLIF) E 5179M 17M 4.74 mJ 0.069   vth 0.3
Ours (SSAM+AiLIF)& E & 7314M &	66M &	6.89 $\mathrm{mJ}$ & 0.092\\   % S3A1
Ours (SSAM+AiLIF)& E+F & 7211M &66M &	6.79 $\mathrm{mJ}$ & 0.091 \\   % S3A2
\bottomrule
\end{tabular}}
\label{energy_ddd17}
\end{table}
\begin{table}[!htbp]
\centering
\caption{The operation number and energy cost of different networks on the DSEC dataset. E and F denote events and images, respectively, for input in inference. \#Add. and \#Mult. denote the number of addition and multiplication operations, respectively. FR denotes the mean firing rate of the model.}
\label{tab:energy_dsec}
 \setlength{\tabcolsep}{1.2mm}{
\begin{tabular}{lrrrrc}
\toprule Method& Input & {\#Add.} &{\#Mult.} & {Energy}& {FR}\\ 
\midrule
ESS (ANN)  &E  & 46850M &46850M & 215.51 $\mathrm{mJ}$ & -\\
\midrule
Ours & E & 38428M &	86M &	35.90 $\mathrm{mJ}$  & 0.125\\   %53.04%
% Ours (LIF) & E & 32277M &	68M &	29.30 $\mathrm{mJ}$  & 0.105\\  %52.71
Ours (SSAM)  &E+F&35332M &	162M& 32.40 $\mathrm{mJ}$  & 0.111\\  % 57.77
% Ours (SSAM+LIF)  &E+F&27912M &	\textcolor{blue}{144M}& 25.17 $\mathrm{mJ}$  & 0.086\\  % 57.22
\bottomrule
\end{tabular}}
\label{energy_dsec}
\end{table}
SNNs contrast with traditional ANNs in their computational approach. 
While ANNs predominantly rely on dense matrix multiplication, SNNs execute event-driven sparse computations, making them potentially more suitable for low-power computing contexts. 
This part delineates the computational costs of our SpikingEDN in comparison to those of conventional ANNs. Fig. \ref{sparsity_hist} showcases the varying degrees of sparse activations in SNNs across different layers. A notable observation is the significant enhancement of event representation in the first stem layer achieved by the SSAM module, which is evident for both types of input. 
In the decoder, there is a gradual reduction in layer sparsity, thus leading to increasingly dense predictions as the activity progresses.

% \textcolor{blue}{
The SNN operates on a principle of MFI. Within the network, information transmission across synapses is binary and primarily involves additive operations, effectively eliminating the need for multiplications. This binary processing, integral
to the SNN, ensures efficient computation and aligns with the concept of MFI. 
However, it is pertinent to highlight that in the initial processing stages of our SpikingEDN, particularly during input data handling, there is a minimal engagement of multiplication operations. 
This is attributed to: a) the input event data, which is derived from SBT and comprises integer values, and b) the floating-point format of image inputs. 
Consequently, the first layer entails a modest level of multiplication.
% }

% \textcolor{blue}{
In alignment with the approaches delineated in \cite{li2021differentiable,rathi2021diet,kim2021beyond}, we quantified the SNN’s addition operations using the formula $s \times T \times A$, where $s$ represents the mean firing rate across the entire test set, $T$ denotes the simulation time step, and $A$ signifies the total count of addition operations in an ANN with an equivalent architectural setup. 
For convolutional operations, the number of additions, denoted as $A_{conv}$, is calculated as follows:
% }
\begin{equation}
A_{conv} = k^2 \times H_{\text {out }} \times W_{\text {out }} \times C_{\text {in }} \times C_{\text {out }} 
\label{eq:FLOP_A_ANN}
\end{equation}
where $k$ denotes the kernel size, $H_{out}$ and $W_{out}$ represent the height and width of the output feature map, while $C_{in}$ and $C_{out}$ indicate the number of input and output channels, respectively. 
The number of multiplication operations in the network is calculated as $T \times A$, with $T$ representing the simulation time step and $A$ the count of multiplication operations. 
For convolution operations, this count is identical to the number of additions.

The computation costs for the DDD17 and DSEC-Semantic datasets are detailed in Tables \ref{energy_ddd17} and \ref{energy_dsec}, respectively. 
Our SpikingEDN demonstrates a remarkably lower number of operations compared to traditional ANNs. 
Notably, using a default threshold value ($u_{\mathrm{th}}=0.5$), the AiLIF neuron reduces the mean firing rate of our network by over 10\% relative to the LIF neuron. 
Specifically, with the same default threshold using an LIF neuron at $u_{th} = 0.5$, or with comparable accuracy using an LIF neuron at $u_{th} = 0.2$ (Section IV-C), the AiLIF neuron demonstrates lower energy consumption and firing rates in both cases.
Table \ref{energy_ddd17} also reveals that with the SSAM module, when event data and images are combined, the network's mean firing rate is lower than with purely event-based inputs, which translates to reduced energy consumption.

For estimating energy consumption, we adhere to \cite{horowitz20141}'s study on $45\mathrm{nm}$ CMOS technology, as utilized in previous research \cite{li2021differentiable,rathi2021diet,kim2021beyond}. 
In our SpikingEDN, an addition operation demands $0.9\mathrm{pJ}$, significantly less than the $4.6\mathrm{pJ}$ required for a multiply-accumulate (MAC) operation in ANNs. 
As a result, SpikingEDN operates at an energy efficiency level that is $6-25$ times greater than that of ANNs. It is essential to note that the evaluation of energy efficiency in SNNs is subject to ongoing research and varies depending on the specific hardware used.
Our methodology provides a basis for direct comparison with ANNs, while alternative methods such as the SynOps model \cite{merolla2014million} represent additional viable options for assessing energy efficiency. 

\subsection{Random Seed Experiments}
To obtain the architecture of our SpikingEDN, we conducted the experiment four times, each with a unique random seed. The most effective architecture was selected based on its performance in the validation set following a brief period of training from scratch. The search spanned 20 epochs, partitioned into an initial five epochs for weight initialization and the subsequent 15 epochs dedicated to bi-level optimization. The optimal architecture identified in this phase was noted at the 1st, 7th, 14th, and 20th epochs.

Post-search, the model underwent retraining with channel expansion over 50 epochs, using a mini-batch size of 2. This process utilized the Adam optimizer with a starting learning rate of $0.001$ and momentum parameters $(0.9, 0.999)$. A Poly learning rate decay strategy was also employed. The search process yielded a significant network performance enhancement, approximately 19\%, comparing the pre-search phase to the completion of 15 epochs of search.

Table \ref{tab:randomseed_mean} illustrates the architecture's initial sensitivity to the seed initialization, which, however, shows consistent optimization throughout the search phase. This gradual improvement in semantic segmentation accuracy can be largely attributed to the layer-level optimization strategy employed. A steady advancement in the architecture's efficacy was observed during the search period, demonstrating the effectiveness of this approach in enhancing model performance.

\begin{table}[!htbp]
\centering
\caption{Search process of the encoder architecture on DDD17. Mean MIoU means the average value of 4 random seeds for the same epoch of 1th, 7th, 14th and 20th , respectively.}
\begin{tabular}{c|cccc}
\toprule Search epochs  & 1 & 7 & 14 & 20 \\ 
\midrule
Mean MIoU \% & 29.78 & 29.90 & 34.96 & 35.13 \\
\bottomrule
\end{tabular}
\label{tab:randomseed_mean}
\end{table}

\subsection{Gray-scale Images}
Event-based sensors excel in capturing high-speed dynamics and managing extreme lighting conditions, such as under or over-exposure, where conventional cameras might struggle to deliver reliable input. 
By contrast, the high-resolution imagery from traditional cameras can enrich event-based data, providing enhanced spatial detail. The impact of incorporating gray-scale images is evident in the improved segmentation performance across various object classes, as shown in Table \ref{tab:iou_compare}. The addition of gray-scale images particularly enhances the delineation and representation of smaller objects, resulting in more precise edge contours. This improvement is most prominent in the classes ‘human’ and ‘vehicle’, which experience significant increases in Intersection over Union (IoU) metrics, by 35.82\% and 28.59\% respectively, when gray-scale images are included. These advancements are visually corroborated by the qualitative comparisons presented in Fig. \ref{fig5}, showcasing the refined segmentation accuracy achieved through the integration of gray-scale imagery.

\begin{table}[!htbp]
\centering
\caption{Comparison of IoU for different categories on DDD17 with and without images, based on the network structure illustrated in Fig. 
\ref{fig2}.
% \ref{searched_ddd17}.
}
\setlength{\tabcolsep}{1.2mm}{
\begin{tabular}{c|cccccc}
\toprule 
Classes & Flat& Background& Object& Vegetation& Human& Vehicle  \\ 
\midrule
IoU(E) \% &75.48& 88.27& 6.41& 48.96& 22.49& 54.38   \\
IoU(E+F) \% & 90.91&  95.01&  33.18& 75.04& 58.31& 82.97 \\
Difference & 15.42 &  6.74 & 26.77& 26.08& 35.82& 28.59 \\
\bottomrule
\end{tabular}}
\label{tab:iou_compare}
\end{table}

\section{Conclusions}\label{sec:conclusion}

In this study, we introduced a novel SNN model specifically tailored for event-based semantic segmentation tasks, dubbed SpikingEDN. 
Specifically, SpikingEDN capitalizes on the innate capabilities of SNNs, which excel in event-driven sparse computation, offering promising avenues for low-power applications.
In comparison with traditional ANNs, SpikingEDN exhibits competitive accuracy with significantly fewer operations and a substantial reduction in energy consumption, making it an ideal candidate for applications where resources and power are limited. Moreover, the study also delved into the empirical analysis of adaptive thresholds in events encoding, which is shown to improve the robustness, accuracy and sparsity of SNN models. 

% \textcolor{blue}{
While the results are encouraging, there are also limitations that warrant further exploration. For example, The searched SNN could be further improved using more sparse designs \cite{shen2023esl}. Our empirical evidence on the advantages of using an adaptive threshold for event encoding could benefit from a deeper theoretical investigation. Moreover, our SpikingEDN is primarily utilized on GPU platforms, and the metrics for energy consumption are based on theoretical estimates. Future research directions could involve deploying SpikingEDN on neuromorphic hardware for empirical energy analysis and extending its practical use in real-world scenarios. 
% } 
% It is acknowledged that further research is required to realize the full potential of these novel architectures and methodologies in an ever-growing field of study.
% \section*{Acknowledgments}
% This should be a simple paragraph before the References to thank those individuals and institutions who have supported your work on this article.
% {\appendix[Proof of the Zonklar Equations]
% Use $\backslash${\tt{appendix}} if you have a single appendix:
% Do not use $\backslash${\tt{section}} anymore after $\backslash${\tt{appendix}}, only $\backslash${\tt{section*}}.
% If you have multiple appendixes use $\backslash${\tt{appendices}} then use $\backslash${\tt{section}} to start each appendix.
% You must declare a $\backslash${\tt{section}} before using any $\backslash${\tt{subsection}} or using $\backslash${\tt{label}} ($\backslash${\tt{appendices}} by itself
%  starts a section numbered zero.)}
%{\appendices
%\section*{Proof of the First Zonklar Equation}
%Appendix one text goes here.
% You can choose not to have a title for an appendix if you want by leaving the argument blank
%\section*{Proof of the Second Zonklar Equation}
%Appendix two text goes here.}
 % argument is your BibTeX string definitions and bibliography database(s)
\bibliography{ref}

% Generated by IEEEtran.bst, version: 1.14 (2015/08/26)
\begin{thebibliography}{10}
\providecommand{\url}[1]{#1}
\csname url@samestyle\endcsname
\providecommand{\newblock}{\relax}
\providecommand{\bibinfo}[2]{#2}
\providecommand{\BIBentrySTDinterwordspacing}{\spaceskip=0pt\relax}
\providecommand{\BIBentryALTinterwordstretchfactor}{4}
\providecommand{\BIBentryALTinterwordspacing}{\spaceskip=\fontdimen2\font plus
\BIBentryALTinterwordstretchfactor\fontdimen3\font minus \fontdimen4\font\relax}
\providecommand{\BIBforeignlanguage}[2]{{%
\expandafter\ifx\csname l@#1\endcsname\relax
\typeout{** WARNING: IEEEtran.bst: No hyphenation pattern has been}%
\typeout{** loaded for the language `#1'. Using the pattern for}%
\typeout{** the default language instead.}%
\else
\language=\csname l@#1\endcsname
\fi
#2}}
\providecommand{\BIBdecl}{\relax}
\BIBdecl

\bibitem{ding2024enhancing}
J.~Ding, Z.~Yu, T.~Huang, and J.~K. Liu, ``Enhancing the robustness of spiking neural networks with stochastic gating mechanisms,'' in \emph{Proceedings of the AAAI Conference on Artificial Intelligence}, vol.~38, no.~1, 2024, pp. 492--502.

\bibitem{shen2024efficient}
J.~Shen, W.~Ni, Q.~Xu, and H.~Tang, ``Efficient spiking neural networks with sparse selective activation for continual learning,'' in \emph{Proceedings of the AAAI Conference on Artificial Intelligence}, vol.~38, no.~1, 2024, pp. 611--619.

\bibitem{shen2021hybridsnn}
J.~Shen, Y.~Zhao, J.~K. Liu, and Y.~Wang, ``Hybridsnn: Combining bio-machine strengths by boosting adaptive spiking neural networks,'' \emph{IEEE Transactions on Neural Networks and Learning Systems}, vol.~34, no.~9, pp. 5841--5855, 2021.

\bibitem{serrano2013128}
T.~Serrano-Gotarredona and B.~Linares-Barranco, ``A 128 $\times$ 128 1.5$\%$ contrast sensitivity 0.9$\%$ fpn 3 $\mu$s latency 4 mw asynchronous frame-free dynamic vision sensor using transimpedance preamplifiers,'' \emph{IEEE Journal of Solid-State Circuits}, vol.~48, no.~3, pp. 827--838, 2013.

\bibitem{brandli2014240}
C.~Brandli, R.~Berner, M.~Yang, S.-C. Liu, and T.~Delbruck, ``A 240 $\times$ 180 130 db 3 $\mu$s latency global shutter spatiotemporal vision sensor,'' \emph{IEEE Journal of Solid-State Circuits}, vol.~49, no.~10, pp. 2333--2341, 2014.

\bibitem{son20174}
B.~Son, Y.~Suh, S.~Kim, H.~Jung, J.-S. Kim, C.~Shin, K.~Park, K.~Lee, J.~Park, J.~Woo \emph{et~al.}, ``4.1 a 640 $\times$ 480 dynamic vision sensor with a 9$\mu$m pixel and 300meps address-event representation,'' in \emph{2017 IEEE International Solid-State Circuits Conference (ISSCC)}.\hskip 1em plus 0.5em minus 0.4em\relax IEEE, 2017, pp. 66--67.

\bibitem{chen2024enhancing}
S.~Chen, J.~Zhang, Y.~Zheng, T.~Huang, and Z.~Yu, ``Enhancing motion deblurring in high-speed scenes with spike streams,'' \emph{Advances in Neural Information Processing Systems}, vol.~36, 2024.

\bibitem{roy2019towards}
K.~Roy, A.~Jaiswal, and P.~Panda, ``Towards spike-based machine intelligence with neuromorphic computing,'' \emph{Nature}, vol. 575, no. 7784, pp. 607--617, 2019.

\bibitem{davies2021advancing}
M.~Davies, A.~Wild, G.~Orchard, Y.~Sandamirskaya, G.~A.~F. Guerra, P.~Joshi, P.~Plank, and S.~R. Risbud, ``Advancing neuromorphic computing with loihi: A survey of results and outlook,'' \emph{Proceedings of the IEEE}, vol. 109, no.~5, pp. 911--934, 2021.

\bibitem{merolla2014million}
P.~A. Merolla, J.~V. Arthur, R.~Alvarez-Icaza, A.~S. Cassidy, J.~Sawada, F.~Akopyan, B.~L. Jackson, N.~Imam, C.~Guo, Y.~Nakamura \emph{et~al.}, ``A million spiking-neuron integrated circuit with a scalable communication network and interface,'' \emph{Science}, vol. 345, no. 6197, pp. 668--673, 2014.

\bibitem{furber2014spinnaker}
S.~B. Furber, F.~Galluppi, S.~Temple, and L.~A. Plana, ``The spinnaker project,'' \emph{Proceedings of the IEEE}, vol. 102, no.~5, pp. 652--665, 2014.

\bibitem{kungl2019accelerated}
A.~F. Kungl, S.~Schmitt, J.~Kl{\"a}hn, P.~M{\"u}ller, A.~Baumbach, D.~Dold, A.~Kugele, E.~M{\"u}ller, C.~Koke, M.~Kleider \emph{et~al.}, ``Accelerated physical emulation of bayesian inference in spiking neural networks,'' \emph{Frontiers in Neuroscience}, p. 1201, 2019.

\bibitem{frenkel2022reckon}
C.~Frenkel and G.~Indiveri, ``Reckon: A {28nm} sub-mm2 task-agnostic spiking recurrent neural network processor enabling on-chip learning over second-long timescales,'' in \emph{2022 IEEE International Solid-State Circuits Conference (ISSCC)}, vol.~65.\hskip 1em plus 0.5em minus 0.4em\relax IEEE, 2022, pp. 1--3.

\bibitem{zhang2023automotive}
H.~Zhang, L.~Leng, K.~Che, Q.~Liu, J.~Cheng, Q.~Guo, J.~Liao, and R.~Cheng, ``Automotive object detection via learning sparse events by temporal dynamics of spiking neurons,'' \emph{arXiv preprint arXiv:2307.12900}, 2023.

\bibitem{alonso2019ev}
I.~Alonso and A.~C. Murillo, ``Ev-segnet: Semantic segmentation for event-based cameras,'' in \emph{Proceedings of the IEEE/CVF Conference on Computer Vision and Pattern Recognition Workshops}, 2019, pp. 0--0.

\bibitem{sun2022ess}
Z.~Sun, N.~Messikommer, D.~Gehrig, and D.~Scaramuzza, ``Ess: Learning event-based semantic segmentation from still images,'' in \emph{European Conference on Computer Vision}.\hskip 1em plus 0.5em minus 0.4em\relax Springer, 2022, pp. 341--357.

\bibitem{yang2023event}
Y.~Yang, L.~Pan, and L.~Liu, ``Event camera data pre-training,'' \emph{arXiv preprint arXiv:2301.01928}, 2023.

\bibitem{ahmed2021deep}
S.~H. Ahmed, H.~W. Jang, S.~N. Uddin, and Y.~J. Jung, ``Deep event stereo leveraged by event-to-image translation,'' in \emph{Proceedings of the AAAI Conference on Artificial Intelligence}, vol.~35, no.~2, 2021, pp. 882--890.

\bibitem{zhang2022discrete}
K.~Zhang, K.~Che, J.~Zhang, J.~Cheng, Z.~Zhang, Q.~Guo, and L.~Leng, ``Discrete time convolution for fast event-based stereo,'' in \emph{Proceedings of the IEEE/CVF Conference on Computer Vision and Pattern Recognition}, 2022, pp. 8676--8686.

\bibitem{Luo_2023_ICCV}
X.~Luo, K.~Luo, A.~Luo, Z.~Wang, P.~Tan, and S.~Liu, ``Learning optical flow from event camera with rendered dataset,'' in \emph{Proceedings of the IEEE/CVF International Conference on Computer Vision (ICCV)}, October 2023, pp. 9847--9857.

\bibitem{Ponghiran_2023_ICCV}
W.~Ponghiran, C.~M. Liyanagedera, and K.~Roy, ``Event-based temporally dense optical flow estimation with sequential learning,'' in \emph{Proceedings of the IEEE/CVF International Conference on Computer Vision (ICCV)}, October 2023, pp. 9827--9836.

\bibitem{Liu_2023_ICCV}
H.~Liu, G.~Chen, S.~Qu, Y.~Zhang, Z.~Li, A.~Knoll, and C.~Jiang, ``Tma: Temporal motion aggregation for event-based optical flow,'' in \emph{Proceedings of the IEEE/CVF International Conference on Computer Vision (ICCV)}, October 2023, pp. 9685--9694.

\bibitem{Su_2023_ICCV}
Q.~Su, Y.~Chou, Y.~Hu, J.~Li, S.~Mei, Z.~Zhang, and G.~Li, ``Deep directly-trained spiking neural networks for object detection,'' in \emph{Proceedings of the IEEE/CVF International Conference on Computer Vision (ICCV)}, October 2023, pp. 6555--6565.

\bibitem{zenke2018superspike}
F.~Zenke and S.~Ganguli, ``Superspike: Supervised learning in multilayer spiking neural networks,'' \emph{Neural Computation}, vol.~30, no.~6, pp. 1514--1541, 2018.

\bibitem{wu2018spatio}
Y.~Wu, L.~Deng, G.~Li, J.~Zhu, and L.~Shi, ``Spatio-temporal backpropagation for training high-performance spiking neural networks,'' \emph{Frontiers in Neuroscience}, vol.~12, p. 331, 2018.

\bibitem{neftci2019surrogate}
E.~O. Neftci, H.~Mostafa, and F.~Zenke, ``Surrogate gradient learning in spiking neural networks: Bringing the power of gradient-based optimization to spiking neural networks,'' \emph{IEEE Signal Processing Magazine}, vol.~36, no.~6, pp. 51--63, 2019.

\bibitem{rathi2021diet}
N.~Rathi and K.~Roy, ``Diet-snn: A low-latency spiking neural network with direct input encoding and leakage and threshold optimization,'' \emph{IEEE Transactions on Neural Networks and Learning Systems}, 2021.

\bibitem{zheng2021going}
H.~Zheng, Y.~Wu, L.~Deng, Y.~Hu, and G.~Li, ``Going deeper with directly-trained larger spiking neural networks,'' in \emph{Proceedings of the AAAI Conference on Artificial Intelligence}, vol.~35, no.~12, 2021, pp. 11\,062--11\,070.

\bibitem{li2021differentiable}
Y.~Li, Y.~Guo, S.~Zhang, S.~Deng, Y.~Hai, and S.~Gu, ``Differentiable spike: Rethinking gradient-descent for training spiking neural networks,'' \emph{Advances in Neural Information Processing Systems}, vol.~34, 2021.

\bibitem{fang2021deep}
W.~Fang, Z.~Yu, Y.~Chen, T.~Huang, T.~Masquelier, and Y.~Tian, ``Deep residual learning in spiking neural networks,'' \emph{Advances in Neural Information Processing Systems}, vol.~34, 2021.

\bibitem{deng2022temporal}
S.~Deng, Y.~Li, S.~Zhang, and S.~Gu, ``Temporal efficient training of spiking neural network via gradient re-weighting,'' \emph{arXiv preprint arXiv:2202.11946}, 2022.

\bibitem{hagenaars2021self}
J.~Hagenaars, F.~Paredes-Vall{\'e}s, and G.~De~Croon, ``Self-supervised learning of event-based optical flow with spiking neural networks,'' \emph{Advances in Neural Information Processing Systems}, vol.~34, 2021.

\bibitem{kim2021beyond}
Y.~Kim, J.~Chough, and P.~Panda, ``Beyond classification: Directly training spiking neural networks for semantic segmentation,'' \emph{arXiv preprint arXiv:2110.07742}, 2021.

\bibitem{zhu2022event}
L.~Zhu, X.~Wang, Y.~Chang, J.~Li, T.~Huang, and Y.~Tian, ``Event-based video reconstruction via potential-assisted spiking neural network,'' \emph{arXiv preprint arXiv:2201.10943}, 2022.

\bibitem{vaswani2017attention}
A.~Vaswani, N.~Shazeer, N.~Parmar, J.~Uszkoreit, L.~Jones, A.~N. Gomez, {\L}.~Kaiser, and I.~Polosukhin, ``Attention is all you need,'' \emph{Advances in Neural Information Processing Systems}, vol.~30, 2017.

\bibitem{ba2016layer}
J.~L. Ba, J.~R. Kiros, and G.~E. Hinton, ``Layer normalization,'' \emph{arXiv preprint arXiv:1607.06450}, 2016.

\bibitem{ulyanov2016instance}
D.~Ulyanov, A.~Vedaldi, and V.~Lempitsky, ``Instance normalization: The missing ingredient for fast stylization,'' \emph{arXiv Preprint arXiv:1607.08022}, 2016.

\bibitem{liu2019auto}
C.~Liu, L.-C. Chen, F.~Schroff, H.~Adam, W.~Hua, A.~L. Yuille, and L.~Fei-Fei, ``{Auto-DeepLab: Hierarchical Neural Architecture Search for Semantic Image Segmentation},'' in \emph{2019 IEEE/CVF Conference on Computer Vision and Pattern Recognition (CVPR)}.\hskip 1em plus 0.5em minus 0.4em\relax IEEE Computer Society, 2019, pp. 82--92.

\bibitem{che2022differentiable}
\BIBentryALTinterwordspacing
K.~Che, L.~Leng, K.~Zhang, J.~Zhang, Q.~Meng, J.~Cheng, Q.~Guo, and J.~Liao, ``Differentiable hierarchical and surrogate gradient search for spiking neural networks,'' in \emph{Advances in Neural Information Processing Systems}, 2022. [Online]. Available: \url{https://openreview.net/forum?id=Lr2Z85cdvB}
\BIBentrySTDinterwordspacing

\bibitem{binas2017ddd17}
J.~Binas, D.~Neil, S.-C. Liu, and T.~Delbruck, ``Ddd17: End-to-end davis driving dataset,'' \emph{arXiv preprint arXiv:1711.01458}, 2017.

\bibitem{gehrig2021dsec}
M.~Gehrig, W.~Aarents, D.~Gehrig, and D.~Scaramuzza, ``Dsec: A stereo event camera dataset for driving scenarios,'' \emph{IEEE Robotics and Automation Letters}, vol.~6, no.~3, pp. 4947--4954, 2021.

\bibitem{chollet2017xception}
F.~Chollet, ``Xception: Deep learning with depthwise separable convolutions,'' in \emph{Proceedings of the IEEE Conference on Computer Vision and Pattern Recognition}, 2017, pp. 1251--1258.

\bibitem{wang2021evdistill}
L.~Wang, Y.~Chae, S.-H. Yoon, T.-K. Kim, and K.-J. Yoon, ``Evdistill: Asynchronous events to end-task learning via bidirectional reconstruction-guided cross-modal knowledge distillation,'' in \emph{Proceedings of the IEEE/CVF Conference on Computer Vision and Pattern Recognition}, 2021, pp. 608--619.

\bibitem{xie2024cross}
C.~Xie, W.~Gao, and R.~Guo, ``Cross-modal learning for event-based semantic segmentation via attention soft alignment,'' \emph{IEEE Robotics and Automation Letters}, 2024.

\bibitem{chen2017deeplab}
L.-C. Chen, G.~Papandreou, I.~Kokkinos, K.~Murphy, and A.~L. Yuille, ``Deeplab: Semantic image segmentation with deep convolutional nets, atrous convolution, and fully connected crfs,'' \emph{IEEE Transactions on Pattern Analysis and Machine Intelligence}, vol.~40, no.~4, pp. 834--848, 2017.

\bibitem{long2015fully}
J.~Long, E.~Shelhamer, and T.~Darrell, ``Fully convolutional networks for semantic segmentation,'' in \emph{Proceedings of the IEEE Conference on Computer Vision and Pattern Recognition}, 2015, pp. 3431--3440.

\bibitem{das2024halsie}
S.~Das~Biswas, A.~Kosta, C.~Liyanagedera, M.~Apolinario, and K.~Roy, ``Halsie: Hybrid approach to learning segmentation by simultaneously exploiting image and event modalities,'' in \emph{Proceedings of the IEEE/CVF Winter Conference on Applications of Computer Vision}, 2024, pp. 5964--5974.

\bibitem{shrestha2018slayer}
S.~B. Shrestha and G.~Orchard, ``Slayer: Spike layer error reassignment in time,'' \emph{Advances in Neural Information Processing Systems}, vol.~31, 2018.

\bibitem{wu2019direct}
Y.~Wu, L.~Deng, G.~Li, J.~Zhu, Y.~Xie, and L.~Shi, ``Direct training for spiking neural networks: Faster, larger, better,'' in \emph{Proceedings of the AAAI Conference on Artificial Intelligence}, vol.~33, no.~01, 2019, pp. 1311--1318.

\bibitem{wozniak2020deep}
S.~Wo{\'z}niak, A.~Pantazi, T.~Bohnstingl, and E.~Eleftheriou, ``Deep learning incorporating biologically inspired neural dynamics and in-memory computing,'' \emph{Nature Machine Intelligence}, vol.~2, no.~6, pp. 325--336, 2020.

\bibitem{dampfhoffer2023backpropagation}
M.~Dampfhoffer, T.~Mesquida, A.~Valentian, and L.~Anghel, ``Backpropagation-based learning techniques for deep spiking neural networks: A survey,'' \emph{IEEE Transactions on Neural Networks and Learning Systems}, 2023.

\bibitem{zhou2022spikformer}
Z.~Zhou, Y.~Zhu, C.~He, Y.~Wang, S.~Yan, Y.~Tian, and L.~Yuan, ``Spikformer: When spiking neural network meets transformer,'' \emph{arXiv preprint arXiv:2209.15425}, 2022.

\bibitem{ranccon2021stereospike}
U.~Ran{\c{c}}on, J.~Cuadrado-Anibarro, B.~R. Cottereau, and T.~Masquelier, ``Stereospike: Depth learning with a spiking neural network,'' \emph{arXiv preprint arXiv:2109.13751}, 2021.

\bibitem{liu2018darts}
H.~Liu, K.~Simonyan, and Y.~Yang, ``{DARTS: Differentiable Architecture Search},'' \emph{arXiv preprint arXiv:1806.09055}, 2018.

\bibitem{bellec2018long}
G.~Bellec, D.~Salaj, A.~Subramoney, R.~Legenstein, and W.~Maass, ``Long short-term memory and learning-to-learn in networks of spiking neurons,'' \emph{Advances in Neural Information Processing Systems}, vol.~31, 2018.

\bibitem{bellec2020solution}
G.~Bellec, F.~Scherr, A.~Subramoney, E.~Hajek, D.~Salaj, R.~Legenstein, and W.~Maass, ``A solution to the learning dilemma for recurrent networks of spiking neurons,'' \emph{Nature Communications}, vol.~11, no.~1, p. 3625, 2020.

\bibitem{wang2019event}
L.~Wang, Y.-S. Ho, K.-J. Yoon \emph{et~al.}, ``Event-based high dynamic range image and very high frame rate video generation using conditional generative adversarial networks,'' in \emph{Proceedings of the IEEE/CVF Conference on Computer Vision and Pattern Recognition}, 2019, pp. 10\,081--10\,090.

\bibitem{fang2021incorporating}
W.~Fang, Z.~Yu, Y.~Chen, T.~Masquelier, T.~Huang, and Y.~Tian, ``Incorporating learnable membrane time constant to enhance learning of spiking neural networks,'' in \emph{Proceedings of the IEEE/CVF International Conference on Computer Vision}, 2021, pp. 2661--2671.

\bibitem{park2019semantic}
T.~Park, M.-Y. Liu, T.-C. Wang, and J.-Y. Zhu, ``Semantic image synthesis with spatially-adaptive normalization,'' in \emph{Proceedings of the IEEE/CVF Conference on Computer Vision and Pattern Recognition}, 2019, pp. 2337--2346.

\bibitem{cadena2021spade}
P.~R.~G. Cadena, Y.~Qian, C.~Wang, and M.~Yang, ``Spade-e2vid: Spatially-adaptive denormalization for event-based video reconstruction,'' \emph{IEEE Transactions on Image Processing}, vol.~30, pp. 2488--2500, 2021.

\bibitem{DBLP:journals/corr/ChenPSA17}
\BIBentryALTinterwordspacing
L.~Chen, G.~Papandreou, F.~Schroff, and H.~Adam, ``Rethinking atrous convolution for semantic image segmentation,'' \emph{CoRR}, vol. abs/1706.05587, 2017. [Online]. Available: \url{http://arxiv.org/abs/1706.05587}
\BIBentrySTDinterwordspacing

\bibitem{ioffe2015batch}
S.~Ioffe and C.~Szegedy, ``Batch normalization: Accelerating deep network training by reducing internal covariate shift,'' in \emph{International Conference on Machine Learning}.\hskip 1em plus 0.5em minus 0.4em\relax PMLR, 2015, pp. 448--456.

\bibitem{sengupta2019going}
A.~Sengupta, Y.~Ye, R.~Wang, C.~Liu, and K.~Roy, ``Going deeper in spiking neural networks: Vgg and residual architectures,'' \emph{Frontiers in Neuroscience}, vol.~13, p.~95, 2019.

\bibitem{hu2021spiking}
Y.~Hu, H.~Tang, and G.~Pan, ``Spiking deep residual networks,'' \emph{IEEE Transactions on Neural Networks and Learning Systems}, 2021.

\bibitem{schnider2023neuromorphic}
Y.~Schnider, S.~Wo{\'z}niak, M.~Gehrig, J.~Lecomte, A.~Von~Arnim, L.~Benini, D.~Scaramuzza, and A.~Pantazi, ``Neuromorphic optical flow and real-time implementation with event cameras,'' in \emph{Proceedings of the IEEE/CVF Conference on Computer Vision and Pattern Recognition}, 2023, pp. 4128--4137.

\bibitem{shen2023esl}
J.~Shen, Q.~Xu, J.~K. Liu, Y.~Wang, G.~Pan, and H.~Tang, ``Esl-snns: An evolutionary structure learning strategy for spiking neural networks,'' in \emph{Proceedings of the AAAI Conference on Artificial Intelligence}, vol.~37, no.~1, 2023, pp. 86--93.

\bibitem{he2016deep}
K.~He, X.~Zhang, S.~Ren, and J.~Sun, ``Deep residual learning for image recognition,'' in \emph{Proceedings of the IEEE Conference on Computer Vision and Pattern Recognition}, 2016, pp. 770--778.

\bibitem{vicente2022keys}
A.~Vicente-Sola, D.~L. Manna, P.~Kirkland, G.~Di~Caterina, and T.~Bihl, ``Keys to accurate feature extraction using residual spiking neural networks,'' \emph{Neuromorphic Computing and Engineering}, vol.~2, no.~4, p. 044001, 2022.

\bibitem{horowitz20141}
M.~Horowitz, ``1.1 computing's energy problem (and what we can do about it),'' in \emph{2014 IEEE International Solid-State Circuits Conference Digest of Technical Papers (ISSCC)}.\hskip 1em plus 0.5em minus 0.4em\relax IEEE, 2014, pp. 10--14.

\end{thebibliography}
%
% \section{Simple References}
% You can manually copy in the resultant .bbl file and set second argument of $\backslash${\tt{begin}} to the number of references
%  (used to reserve space for the reference number labels box).
% \begin{thebibliography}{1}
\bibliographystyle{IEEEtran}
% \bibitem{ref1}
% {\it{Mathematics Into Type}}. American Mathematical Society. [Online]. Available: https://www.ams.org/arc/styleguide/mit-2.pdf
% \bibitem{ref2}
% T. W. Chaundy, P. R. Barrett and C. Batey, {\it{The Printing of Mathematics}}. London, U.K., Oxford Univ. Press, 1954.

% \bibitem{ref3}
% F. Mittelbach and M. Goossens, {\it{The \LaTeX Companion}}, 2nd ed. Boston, MA, USA: Pearson, 2004.

% \bibitem{ref4}
% G. Gr\"atzer, {\it{More Math Into LaTeX}}, New York, NY, USA: Springer, 2007.

% \bibitem{ref5}M. Letourneau and J. W. Sharp, {\it{AMS-StyleGuide-online.pdf,}} American Mathematical Society, Providence, RI, USA, [Online]. Available: http://www.ams.org/arc/styleguide/index.html

% \bibitem{ref6}
% H. Sira-Ramirez, ``On the sliding mode control of nonlinear systems,'' \textit{Syst. Control Lett.}, vol. 19, pp. 303--312, 1992.

% \bibitem{ref7}
% A. Levant, ``Exact differentiation of signals with unbounded higher derivatives,''  in \textit{Proc. 45th IEEE Conf. Decis.
% Control}, San Diego, CA, USA, 2006, pp. 5585--5590. DOI: 10.1109/CDC.2006.377165.

% \bibitem{ref8}
% M. Fliess, C. Join, and H. Sira-Ramirez, ``Non-linear estimation is easy,'' \textit{Int. J. Model., Ident. Control}, vol. 4, no. 1, pp. 12--27, 2008.

% \bibitem{ref9}
% R. Ortega, A. Astolfi, G. Bastin, and H. Rodriguez, ``Stabilization of food-chain systems using a port-controlled Hamiltonian description,'' in \textit{Proc. Amer. Control Conf.}, Chicago, IL, USA,
% 2000, pp. 2245--2249.

% \end{thebibliography}

% bio

\begin{IEEEbiography}[{\includegraphics[width=1in,height=1.2in,clip,keepaspectratio]{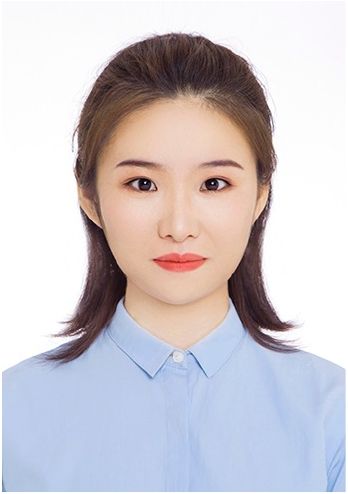}}]{Rui Zhang} received the Master degree from Southern University of Science and Technology, China, in 2023, and the B.Eng. degree from the Northeastern University, China, in 2018. Her research interests include Computer Vision and Spiking Neural Networks.
\end{IEEEbiography}

\begin{IEEEbiography}[{\includegraphics[width=1in,height=1.2in,clip,keepaspectratio]{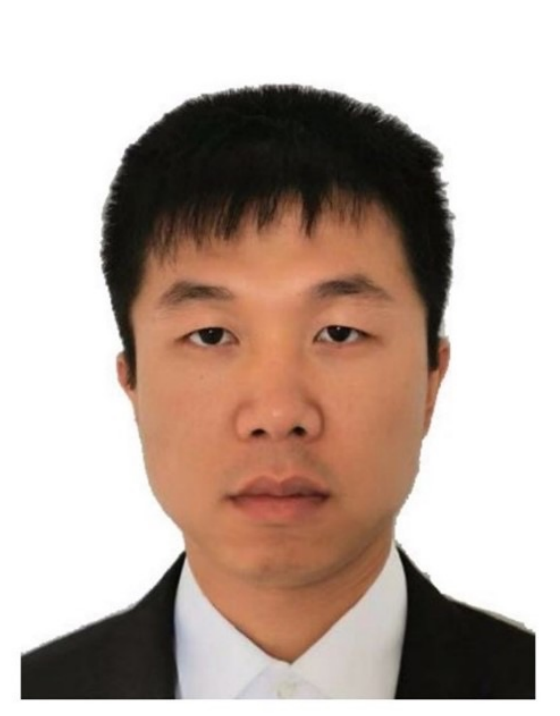}}]{Luziwei Leng} (Member, IEEE) received the B.Sc. degree from the University of Electronic Science and Technology of China, Chengdu, China, in 2011, and the Master and the Doctoral degree from the Department of Physics, Heidelberg University, Germany in 2014 and 2019, respectively. He is currently a Principal Engineer at the Advanced Computing and Storage Lab of Central Research Institute, Huawei Technologies Co., Ltd and an Industry Supervisor of the joint-master program between Huawei and the Southern University of Science and Technology, Shenzhen, China. His research focuses on brain-inspired computing and machine learning, including spiking neural networks, event-based vision, bio-inspired learning rules and evolutionary algorithms, etc.
\end{IEEEbiography}

\vspace{7em}

\begin{IEEEbiography}[{\includegraphics[width=1in, height=1.2in, clip, keepaspectratio]{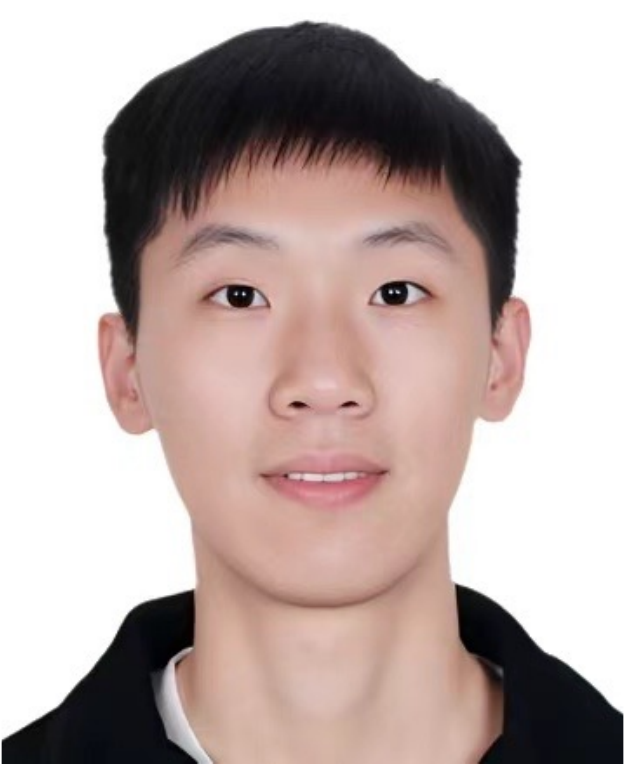}}]{Kaiwei Che} received the B.Eng. degree from Shenzhen University, in 2020, and the M.Eng. degree from the Southern University of Science and Technology, in 2023. He is currently a Ph.D. student at Peking University. His research interests include deep learning and brain-inspired algorithms.
\end{IEEEbiography}

\begin{IEEEbiography}[{\includegraphics[width=1in, height=1.2in, clip, keepaspectratio]{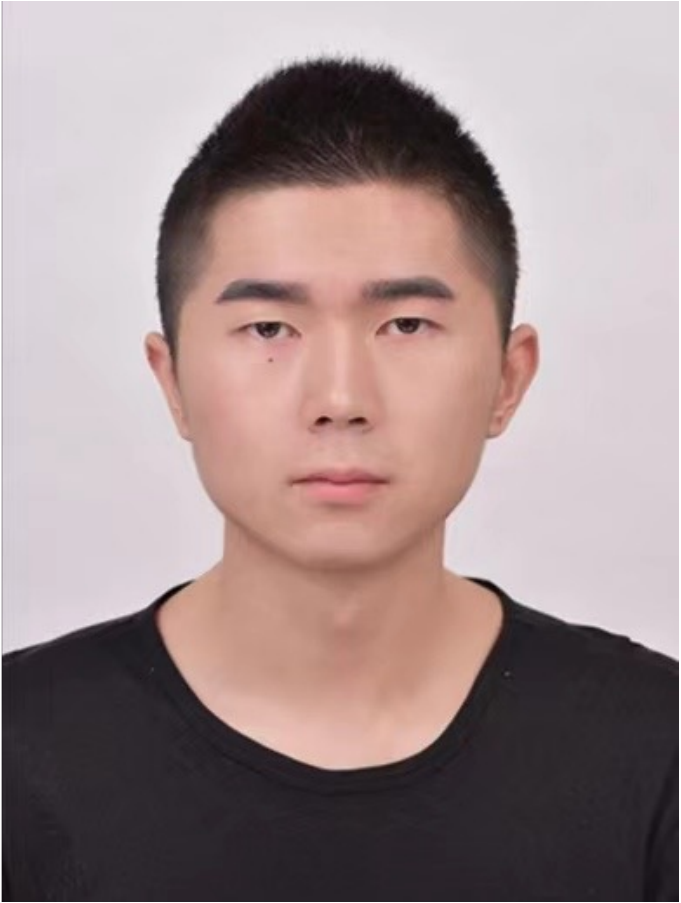}}]{Hu Zhang} received the B.Eng. degree from Huazhong University of Science and Technology, Wuhan, China, in 2018, and the M.Eng. degree from Southern University of Science and Technology, Shenzhen, China, in 2023. He is currently working at BYD Co., Ltd. as a performance development engineer. His research interests include computer vision, deep learning, spiking neural networks and intelligent driving.
\end{IEEEbiography}

\begin{IEEEbiography}[{\includegraphics[width=1in, height=1.2in, clip, keepaspectratio]{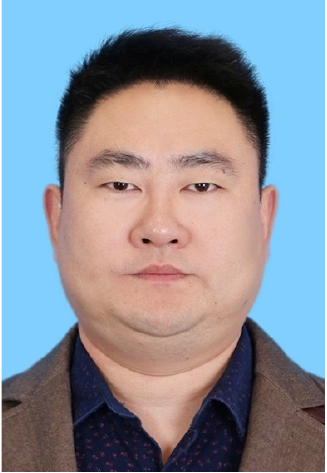}}]{Jie Cheng} received the Ph.D. degree from the Huazhong University of Science and Technology in 2011. He worked as a postdoctoral researcher in Prince Edward Island University from 2011 to 2015. He is currently a Senior Staff Engineer with the Application Innovation Lab, Huawei Technologies Company, Ltd., China. His research focuses on brain-inspired computing, artificial Intelligence and machine learning, including spiking neural neworks, large language model, artificial Intelligence for science, etc. 
\end{IEEEbiography}

\begin{IEEEbiography}[{\includegraphics[width=1in, height=1.2in, clip, keepaspectratio]{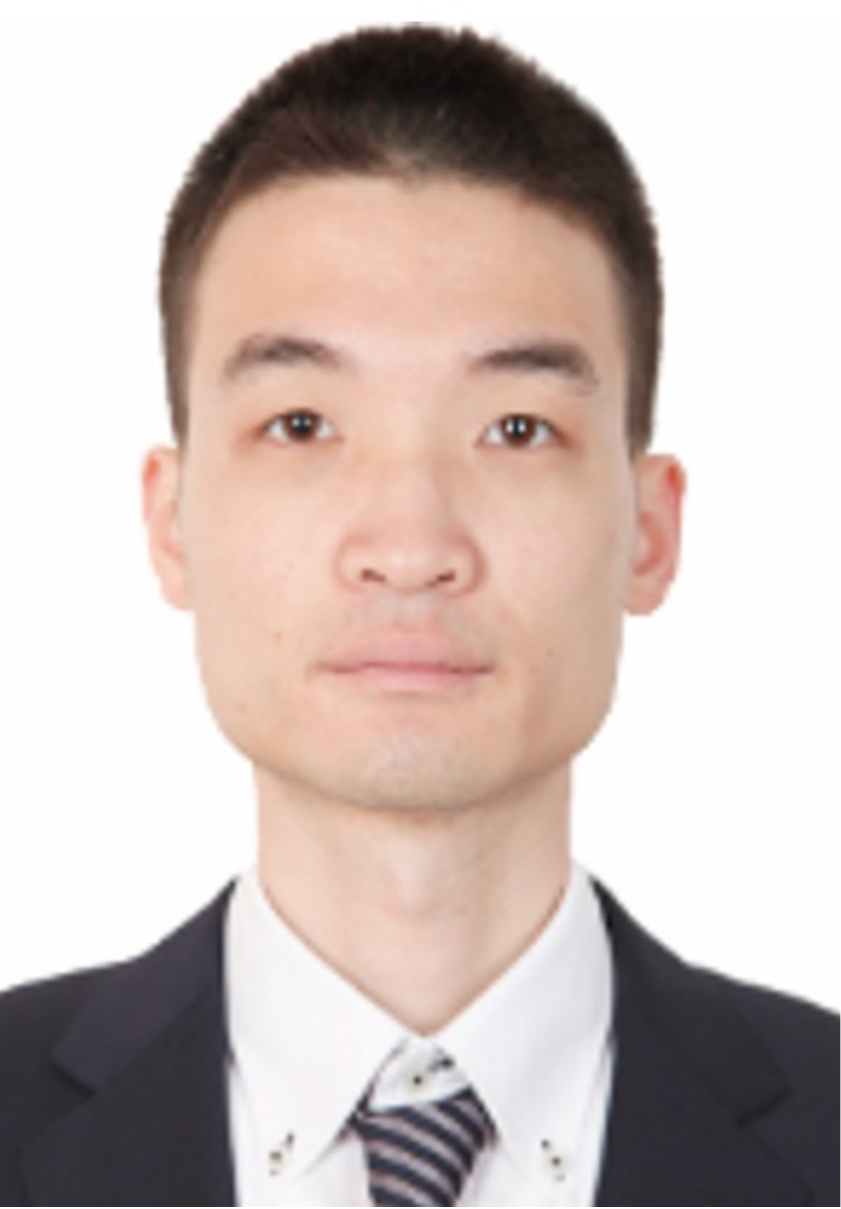}}]{Qinghai Guo} is currently a Principal Engineer at the Advanced Computing and Storage Lab of Huawei Technologies Co., Ltd. and the Team Leader of the Brain-inspired Computing Team. He received his Doctoral degree from the Institute of Mathematical Stochastics, University of Goettingen, Germany in 2017. His research focuses on brain-inspired computing and machine learning, including spiking neural networks, biological plausible and efficient learning theories, neuromorphic computing architecture, etc. 
\end{IEEEbiography}

\begin{IEEEbiography}[{\includegraphics[width=1in,height=1.2in,clip,keepaspectratio]{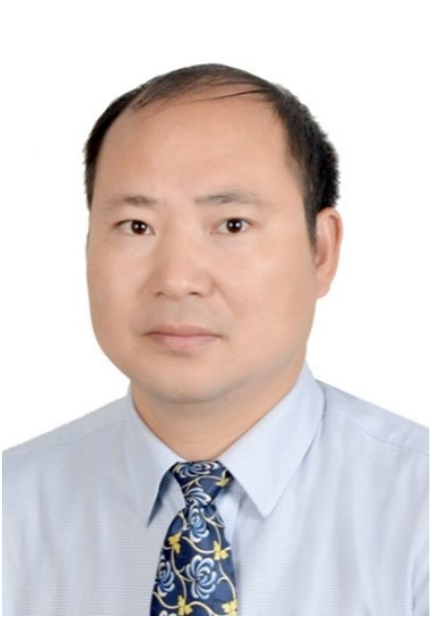}}]{Jianxing Liao} received the Bachelor and the Master degree from Xidian University, Xi’an, China, in 1997 and 2000, respectively. He is currently an Expert at the Advanced Computing and Storage Lab of Central Research Institute, Huawei Technologies Co., Ltd and an Industry Supervisor of the joint-master program between Huawei and the Southern University of Science and Technology, Shenzhen, China. His research focuses on brain-inspired computing and neuromorphic chips, including event-based vision, brain-computer interface, etc.
\end{IEEEbiography}

\begin{IEEEbiography}[{\includegraphics[width=1in,height=1.2in,clip,keepaspectratio]{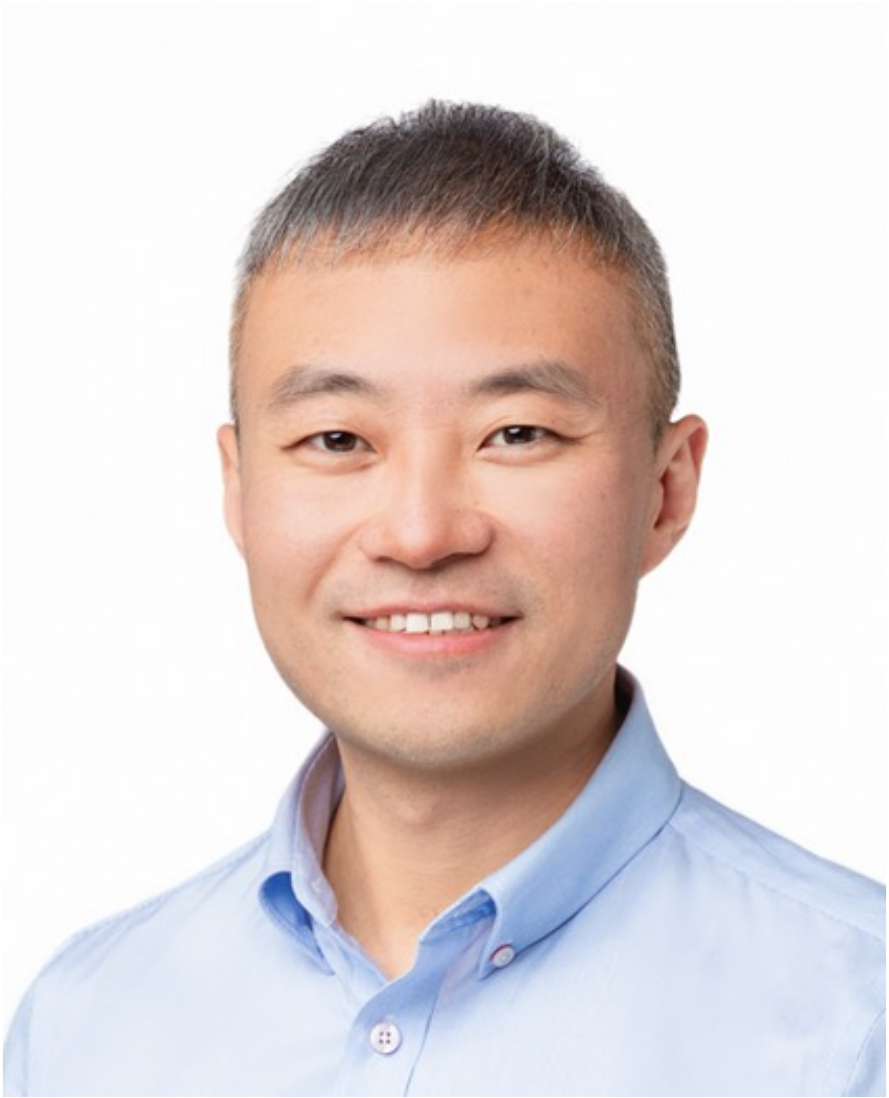}}]{Ran Cheng} (Senior Member, IEEE)
received the B.Sc. degree from the Northeastern University, Shenyang, China, in 2010, and the Ph.D. degree from the University of Surrey, Guildford, U.K., in 2016. He is currently an Associate Professor with the Department of Computer Science and Engineering, Southern University of Science and Technology, Shenzhen, China. He is a recipient of the 2018 and 2021 IEEE Transactions on Evolutionary Computation Outstanding Paper Award, the 2019 IEEE Computational Intelligence Society (CIS) Outstanding Ph.D. Dissertation Award, and the 2020 IEEE Computational Intelligence Magazine Outstanding Paper Award. He is the founding Chair of the IEEE CIS Shenzhen Chapter. He is an Associate Editor of IEEE TEVC, IEEE TAI, IEEE TETCI, and IEEE TCDS.
\end{IEEEbiography}

\end{document}